\documentclass[]{style}

\definecolor{lastauthor}{RGB}{143, 68, 115}

\usepackage{marvosym}
\usepackage{eso-pic}

\title{AutoDesign: Meta-Harness Optimization for Long-Horizon Agentic Design}
\titlelogo{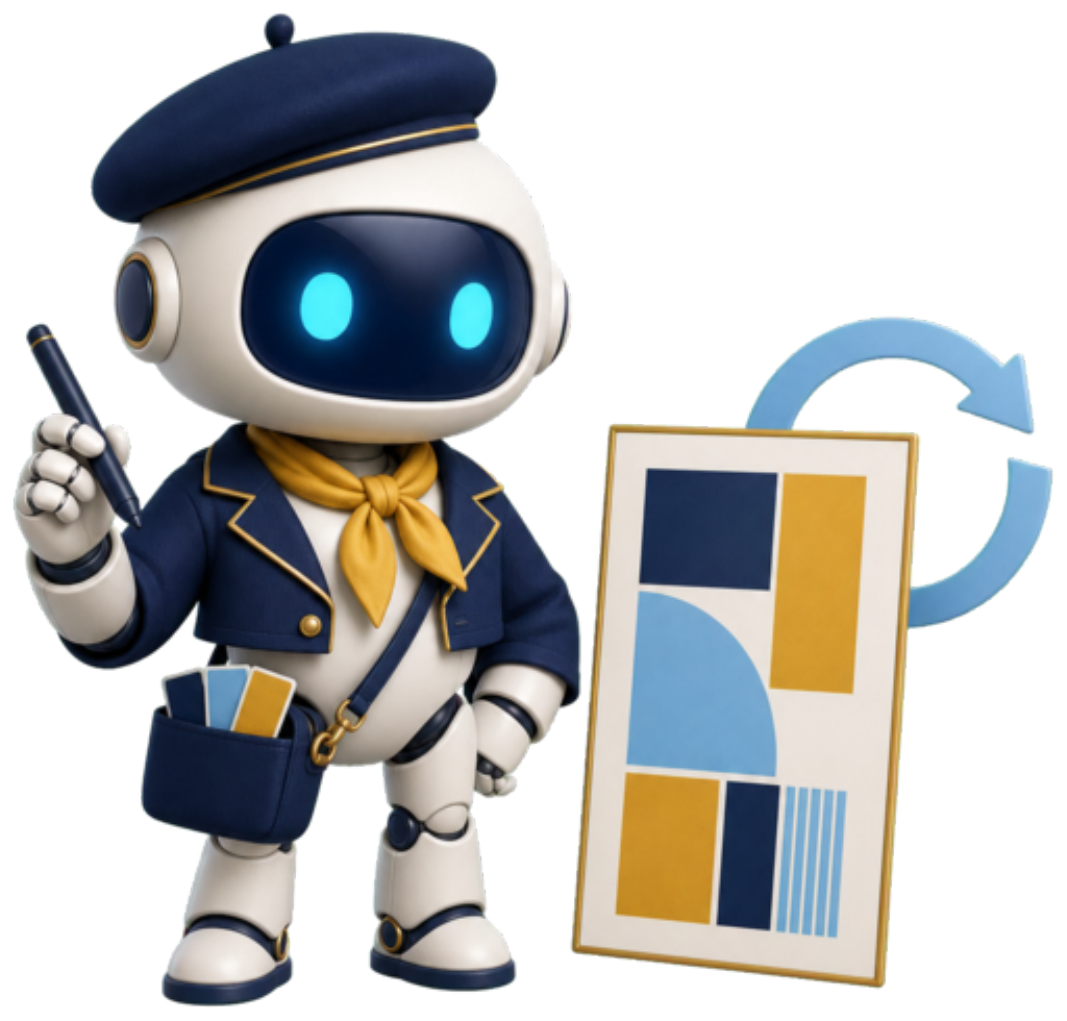}

\author[1,2*]{{Yaxin Luo}}
\author[1*]{{Haobin Jiang}}
\author[1,3]{{Jialv Zou}}
\author[1,4]{{Xu Huang}}
\author[1,5]{{Wenhao Yan}}
\author[1,6]{{Haodong Li}}
\author[1,7]{{Zhengrong Yue}}
\author[1]{{Jing Li}}
\author[2]{{Xiaofu Chen}}
\author[2]{{Xiaohan Zhao}}
\author[2]{{Jiacheng Liu}}
\author[2]{{Jiacheng Cui}}
\author[2\protect\mbox{\protect\Letter}]{{Zhiqiang Shen}}
\author[1\protect\mbox{\protect\Letter},\textdagger]{{Xiaotong Li}}

\affiliation[1]{Meituan}
\affiliation[2]{MBZUAI}
\affiliation[3]{Huazhong University of Science and Technology}
\affiliation[4]{Peking University}
\affiliation[5]{Tsinghua University}
\affiliation[6]{The Chinese University of Hong Kong}
\affiliation[7]{Shanghai Jiao Tong University}
\contribution[*]{Equal contribution}
\contribution[\textdagger]{Project lead}

\usepackage[table]{xcolor}

\usepackage{amsmath}
\usepackage{amssymb}
\usepackage{graphicx}
\usepackage{booktabs}
\usepackage{xcolor}
\usepackage{hyperref}
\usepackage{cleveref}
\usepackage{natbib}
\usepackage{enumitem}
\usepackage{multirow}
\usepackage{subcaption}
\usepackage{microtype}
\usepackage{tabularx}

\usepackage{makecell}
\newcommand{\ie}{\textit{i.e.}}
\newcommand{\eg}{\textit{e.g.}}

\graphicspath{{resources/figures/}}

\hypersetup{
  colorlinks=true,
  linkcolor=metablue,
  citecolor=metablue,
  urlcolor=metablue,
}

\abstract{ 
Transforming multimodal sources into condensed and structured media outputs can be fundamentally conceptualized as a long-horizon agentic process centered on a model-harness system. 
While an ideal harness system should align with human design priors and accumulate reusable experience through empirical exploration to drive recursive self-improvement, existing paradigms remain static and fall short of this capability.
In this paper, we present \texttt{AutoDesign}, a framework that aligns with human design priors, where a meta-harness optimizer guides a code agent to recursively improve harness based on rollout feedback. To instantiate and evaluate this framework, we focus on the academic paper-to-poster generation task and introduce \texttt{PosterBench}, comprising a 100-paper Main Track spanning five disciplines and \texttt{PosterBench-mini}, a shared 10-paper subset for controlled evaluation. On the \texttt{PosterBench} Main Track, \texttt{AutoDesign} achieves the highest score of 78.32, surpassing the closed-source commercial system Claude Design by 7.45 points. Across seven controlled code agent--model configurations, integrating the learned \texttt{DesignHarness} consistently improves performance, increasing the average \texttt{PosterBench} Score from 54.99 to 67.39 (+12.4\%). In a fully autonomous long-horizon loop, it executes 253 tool calls and 11 editing turns within 40 minutes for under \$3, reaching average conference-poster quality in human evaluation. A system-blind human study further demonstrates that \texttt{AutoDesign} achieves the highest human preference among evaluated systems.
}

\newcommand{\autodesignresourceicon}{\includegraphics[width=1.35em]{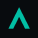}}
\newcommand{\githubresourceicon}{\includegraphics[width=1.35em]{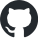}}
\newcommand{\demopageresourceicon}{\includegraphics[width=1.35em]{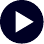}}
\resource{\autodesignresourceicon}{Project Website}{https://autodesign.designanything.ai/}
\resource{\githubresourceicon}{Code Repository}{https://github.com/Yaxin9Luo/AutoDesign}
\resource{\demopageresourceicon}{Demo Page}{https://designanything.ai/}
\newcommand{\firstpagefootnote}{%
  \AddToShipoutPictureFG*{%
    \AtTextLowerLeft{%
      \raisebox{8pt}[0pt][0pt]{%
        \parbox[t]{\textwidth}{\footnotesize
          \rule{0.43\textwidth}{0.35pt}\par\vspace{0.12em}%
          \textbf{Correspondence:} \protect\Letter{}\hspace{0.16em}\email{lixiaotong.edu@gmail.com}\hspace{0.9em}\protect\Letter{}\hspace{0.16em}\email{zhiqiang.shen@mbzuai.ac.ae}\par
          Work done during Yaxin's internship at Meituan.}%
      }%
    }%
  }%
}

\usepackage[dvipsnames]{xcolor}
\definecolor{GainColor}{named}{RawSienna}
\definecolor{BadColor}{named}{Red}

\usepackage{arydshln}
\usepackage[table]{xcolor}  

\newtcolorbox{appendixpromptbox}[1]{
  enhanced,
  breakable,
  colback=metabg!65!white,
  colframe=metablue!72!black,
  colbacktitle=metablue!14!white,
  coltitle=metafg,
  fonttitle=\small\sffamily\bfseries,
  fontupper=\footnotesize\ttfamily,
  title={#1},
  boxrule=0.5pt,
  arc=3pt,
  left=5pt,
  right=5pt,
  top=4pt,
  bottom=4pt,
  before skip=5pt,
  after skip=6pt,
}
\newtcolorbox{appendixrecordbox}[1]{
  enhanced,
  breakable,
  colback=meituanyellow!16!white,
  colframe=meituanlink!82!black,
  colbacktitle=meituanyellow!34!white,
  coltitle=metafg,
  fonttitle=\small\sffamily\bfseries,
  fontupper=\footnotesize,
  title={#1},
  boxrule=0.5pt,
  arc=3pt,
  left=5pt,
  right=5pt,
  top=4pt,
  bottom=4pt,
  before skip=5pt,
  after skip=6pt,
}

\usepackage{epigraph}
\usepackage[normalem]{ulem}
\definecolor{quotemark}{gray}{0.7}
\makeatletter
\def\fquote{%
    \@ifnextchar[{\fquote@i}{\fquote@i[]}
           }%
\def\fquote@i[#1]{%
    \def\tempa{#1}%
    \@ifnextchar[{\fquote@ii}{\fquote@ii[]}
                 }%
\def\fquote@ii[#1]{%
    \def\tempb{#1}%
    \@ifnextchar[{\fquote@iii}{\fquote@iii[]}
                      }%
\def\fquote@iii[#1]{%
    \def\tempc{#1}%
    \vspace{1em}%
    \noindent%
    \begin{list}{}{%
         \setlength{\leftmargin}{0.05\textwidth}%
         \setlength{\rightmargin}{0.05\textwidth}%
                  }%
         \item[]%
         \begin{picture}(0,0)%
         \put(-8,-5){\makebox(0,0){\scalebox{2}{\textcolor{quotemark}{``}}}}%
         \end{picture}%
         \begingroup\itshape}%
 \def\endfquote{%
 \endgroup\par%
 \makebox[0pt][l]{%
 \hspace{0.27\textwidth}%
 \begin{picture}(0,0)(0,0)%
 \put(60,20){\makebox(0,0){%
 \scalebox{2}{\color{quotemark}''}}}%
 \end{picture}}%
 \ifx\tempa\empty%
 \else%
    \ifx\tempc\empty%
       \hfill\rule{110pt}{0.5pt}\\\mbox{}\hfill\tempa,\ \emph{\tempb}%
   \else%
       \hfill\rule{100pt}{0.5pt}\\\mbox{}\hfill\tempa,\ \emph{\tempb},\ \tempc%
   \fi\fi\par%
   \vspace{0.5em}%
 \end{list}%
 }%
 \makeatother

\definecolor{orange}{RGB}{178,92,35}
\definecolor{green1}{RGB}{95,145,51}

\definecolor{red1}{RGB}{197,64,57}
\definecolor{blue1}{RGB}{59,130,220}
\definecolor{green2}{RGB}{82,181,150}
\definecolor{purple1}{RGB}{105,93,223}
\definecolor{orange1}{RGB}{164, 47, 54}
\definecolor{green3}{RGB}{94,145,51}

\definecolor{c_step1}{RGB}{60, 105, 199}
\definecolor{c_step2}{RGB}{159,37,28}
\definecolor{c_step3}{RGB}{107,40,157}

\newenvironment{talign*}
{\csname align*\endcsname}
{\endalign}

\usepackage[utf8]{inputenc}         %
\usepackage[T1]{fontenc}            %
\usepackage{url}                    %
\usepackage{booktabs}               %
\usepackage{amsfonts}               %
\usepackage{nicefrac}               %
\usepackage{microtype}              %
\usepackage{algorithm}
\usepackage{algorithmic}
\usepackage{graphicx}
\usepackage{subcaption}
\usepackage[flushleft]{threeparttable}
\usepackage{float}
\usepackage{multirow}
\usepackage{xspace}
\usepackage{enumitem}
\usepackage[font=small]{caption}
\usepackage{autobreak}
\usepackage{sidecap}
\usepackage{wrapfig}
\usepackage[toc, page, header]{appendix}
\usepackage{tikz}
\usepackage{pifont}
\usepackage{mdframed}
\usepackage{colortbl}

\definecolor{coral}{RGB}{255,127,80}
\definecolor{darkgreen}{RGB}{0,100,0}
\definecolor{darkyellow}{RGB}{204,153,0}
\definecolor{salmon}{RGB}{250,128,114}
\definecolor{darkred}{RGB}{150,0,0}
\newcommand{\darkredtext}[1]{{\color{darkred}#1}}

\newcommand{\secref}[1]{\hyperref[#1]{\darkredtext{Sec.~\ref*{#1}}}}
\newcommand{\thmref}[1]{\hyperref[#1]{\darkredtext{Thm.~\ref*{#1}}}}
\newcommand{\defref}[1]{\hyperref[#1]{\darkredtext{Def.~\ref*{#1}}}}
\newcommand{\propref}[1]{\hyperref[#1]{\darkredtext{Prop.~\ref*{#1}}}}
\newcommand{\assumpref}[1]{\hyperref[#1]{\darkredtext{Assump.~\ref*{#1}}}}
\newcommand{\remarkref}[1]{\hyperref[#1]{\darkredtext{Rem.~\ref*{#1}}}}
\newcommand{\hypref}[1]{\hyperref[#1]{\darkredtext{Hyp.~\ref*{#1}}}}
\newcommand{\conjref}[1]{\hyperref[#1]{\darkredtext{Conj.~\ref*{#1}}}}
\newcommand{\lemref}[1]{\hyperref[#1]{\darkredtext{Lem.~\ref*{#1}}}}
\newcommand{\corref}[1]{\hyperref[#1]{\darkredtext{Cor.~\ref*{#1}}}}
\newcommand{\noteref}[1]{\hyperref[#1]{\darkredtext{Nota.~\ref*{#1}}}}
\newcommand{\claimref}[1]{\hyperref[#1]{\darkredtext{Clm.~\ref*{#1}}}}
\newcommand{\obsref}[1]{\hyperref[#1]{\darkredtext{Obs.~\ref*{#1}}}}
\newcommand{\algref}[1]{\hyperref[#1]{\darkredtext{Alg.~\ref*{#1}}}}
\newcommand{\figref}[1]{\hyperref[#1]{\darkredtext{Figure~\ref*{#1}}}}
\newcommand{\tabref}[1]{\hyperref[#1]{\darkredtext{Table~\ref*{#1}}}}
\newcommand{\appref}[1]{\hyperref[#1]{\darkredtext{App.~\ref*{#1}}}}
\renewcommand{\eqref}[1]{\hyperref[#1]{\darkredtext{Eq.~\ref*{#1}}}}

\usepackage{listings}
\usepackage{xcolor}
\usepackage{booktabs}
\usepackage{amssymb}      
\usepackage{tikz}         

\lstdefinestyle{agentloop}{
  basicstyle=\ttfamily\scriptsize,
  commentstyle=\color{gray!70!black}\itshape,
  keywordstyle=\color{blue!55!black}\bfseries,
  showstringspaces=false,
  columns=fullflexible,
  keepspaces=true,
  escapeinside={(*@}{@*)},
  morekeywords={while,if,not,is,continue,and,or,True,False,None,def,return},
  frame=tb,
  framerule=0.5pt,
  rulecolor=\color{gray!60},
  framesep=5pt,
  backgroundcolor=\color{gray!4},
  xleftmargin=8pt,
  xrightmargin=4pt,
  aboveskip=2pt,
  belowskip=2pt,
  breaklines=false,
}

\definecolor{darkred}{RGB}{165,42,42}

\begin{document}

\maketitle
\firstpagefootnote

\vspace{-0.5em}

\begin{figure}[H]
    \centering
    \newlength{\figureonepanelheight}
    \setlength{\figureonepanelheight}{0.35\textwidth}
    \begin{subfigure}[t]{0.533\textwidth}
        \centering
        \includegraphics[pagebox=cropbox,height=\figureonepanelheight]{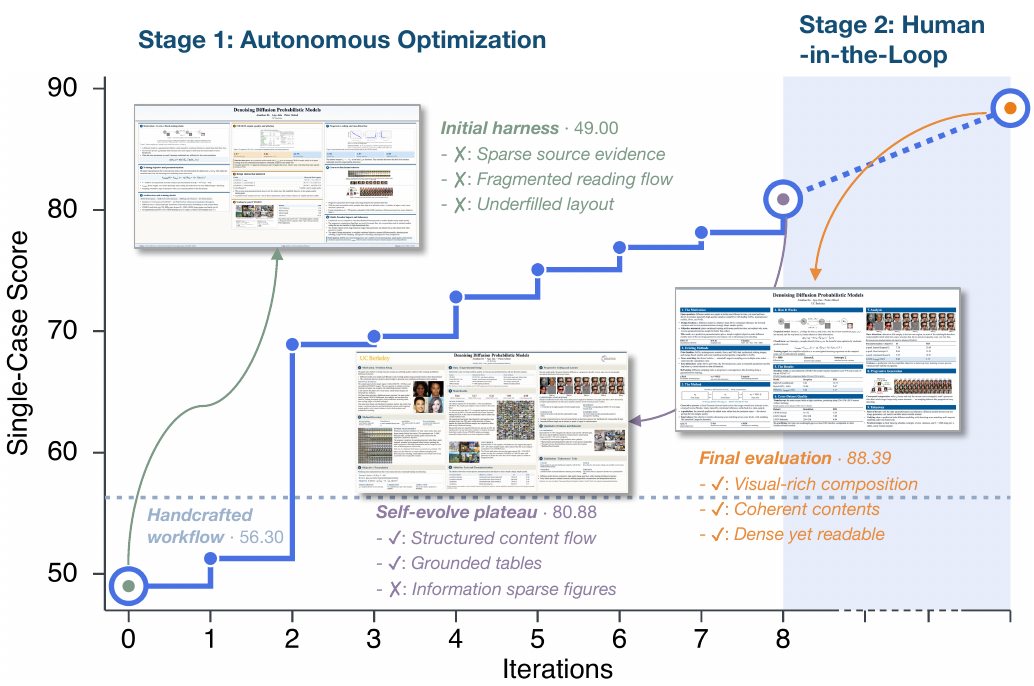}
        \caption{Meta-harness optimization trace}
        \label{fig:autodesign-evolve}
    \end{subfigure}%
    \begin{subfigure}[t]{0.467\textwidth}
        \centering
        \includegraphics[pagebox=cropbox,height=\figureonepanelheight]{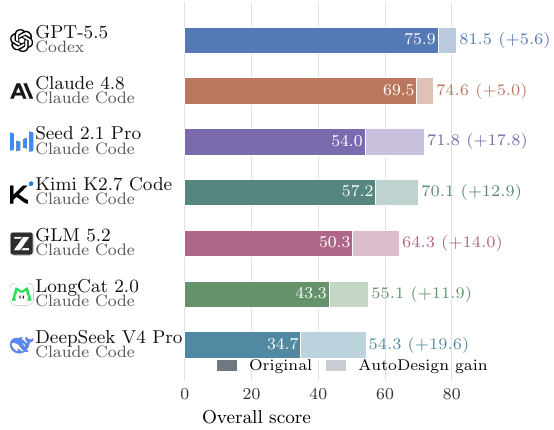}
        \caption{Performance gains from \texttt{DesignHarness}}
        \label{fig:autodesign-benchmark}
    \end{subfigure}
    \caption{\textbf{AutoDesign progressively improves the design harness and the quality of the artifacts.} (a) Score of the poster generated by the design harness for one representative paper, tracked across meta-harness iterations. Autonomous optimization improves the initial harness before reaching a plateau, after which human guidance redirects the search and yields a further gain.
    (b) The optimized harness, \texttt{DesignHarness}, improves all Coding Agents on \texttt{PosterBench} by 5.0 to 19.6 points, achieving a best overall score of 81.5.
    }
    \label{fig:combined}
\end{figure}

\clearpage
\begin{figure}[H]
  \centering
  \includegraphics[width=1.0\textwidth]{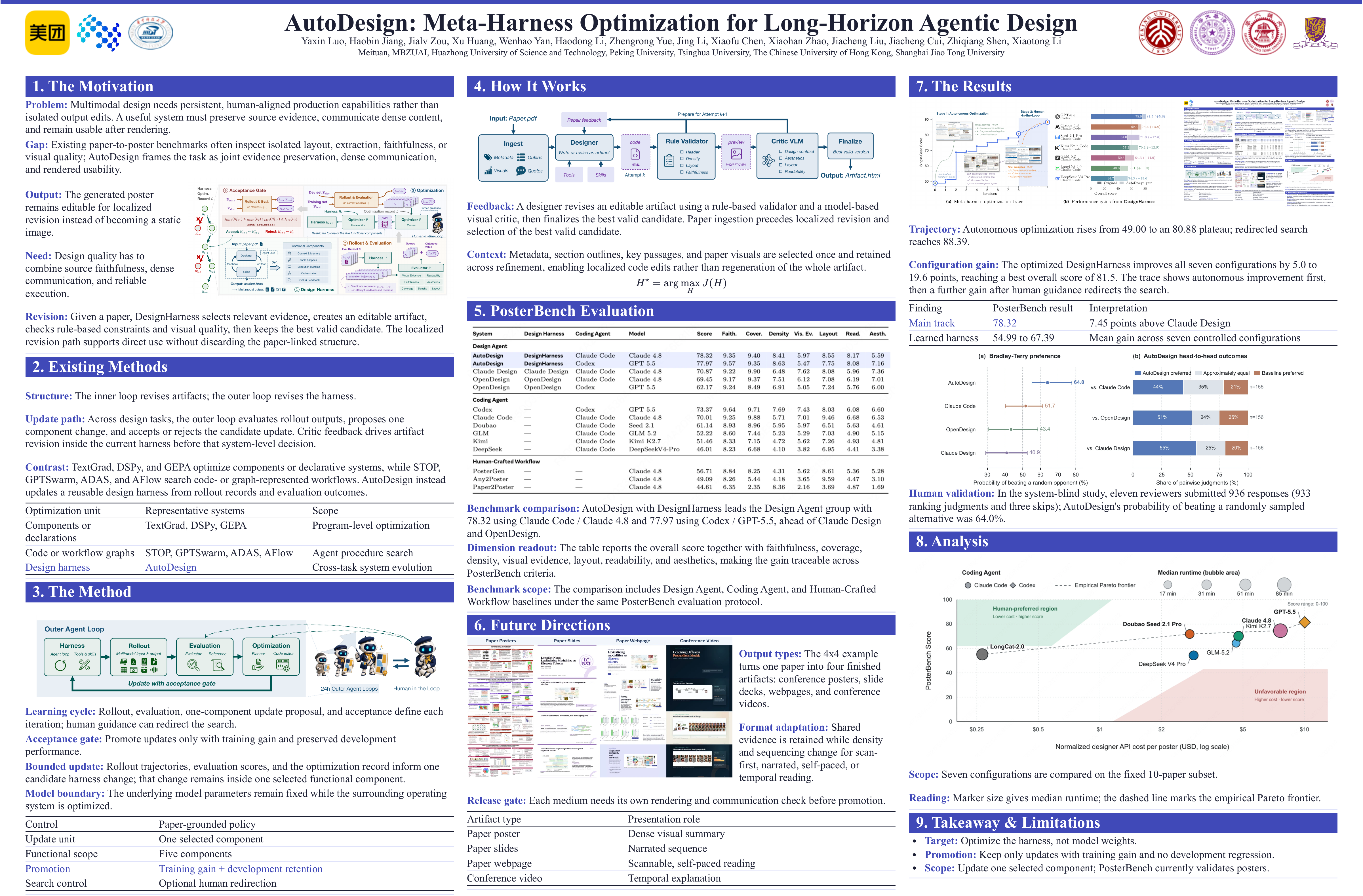}
  \caption{\textbf{AutoDesign for AutoDesign.}
A poster generated by \texttt{AutoDesign} for its own paper, crafted through a long-horizon agentic workflow that autonomously performs source ingestion, iterative generation and refinement, critic feedback integration, and finalization, completing  design process in approximately 40 minutes with negligible human intervention.}
  \label{fig:autodesign-poster-showcase}
\end{figure}

\section{Introduction}
\label{sec:intro}

Human communication often involves understanding, organizing, and presenting information from diverse multimodal sources into human-facing artifacts, \textit{e.g.,} webpages, slides, posters, and videos~\citep{fu2022doc2ppt,pang2025paper2poster,zheng2025pptagent,chen2025paper2web,zhu2025paper2video}.
Achieving such multimodal input-to-output transformation~\citep{wu2023nextgpt,chameleon2024mixed,kim2026dyninomni,longcatnext2026} requires the ability to extract relevant evidence, reason over heterogeneous information, plan intermediate steps, and iteratively improve outputs based on feedback~\citep{choi2026posterforest,liu2026designascode}, which naturally positions multimodal design as a suitable yet challenging long-horizon task for agentic coding. However, developing such systems remains difficult due to the complexity of real-world workflows and the reliance on extensive human feedback.

Current multimodal design systems seek human-aligned design priors through cycles of generation, critique, and revision, often using visual references or design-specific feedback \citep{sun2025p2p,zheng2025pptagent,liu2026designascode}.
At the response level, feedback can revise the current output \citep{madaan2023selfrefine}, while agentic systems can retain reflections, skills, or task experience across attempts \citep{shinn2023reflexion,wang2023voyager,zhao2024expel}. However, unlike human creators who continuously accumulate knowledge from successful revisions and failures, such systems treat individual human-aligned feedback as transient signals rather than reusable design knowledge. The unresolved question for multimodal design is how to convert multimodal evidence, structural constraints, feedback, and human preferences into persistent design-aligned capabilities of the production system \citep{ren2026selfimprovement}.

\begin{figure}[t]
  \centering
  \includegraphics[width=\textwidth]{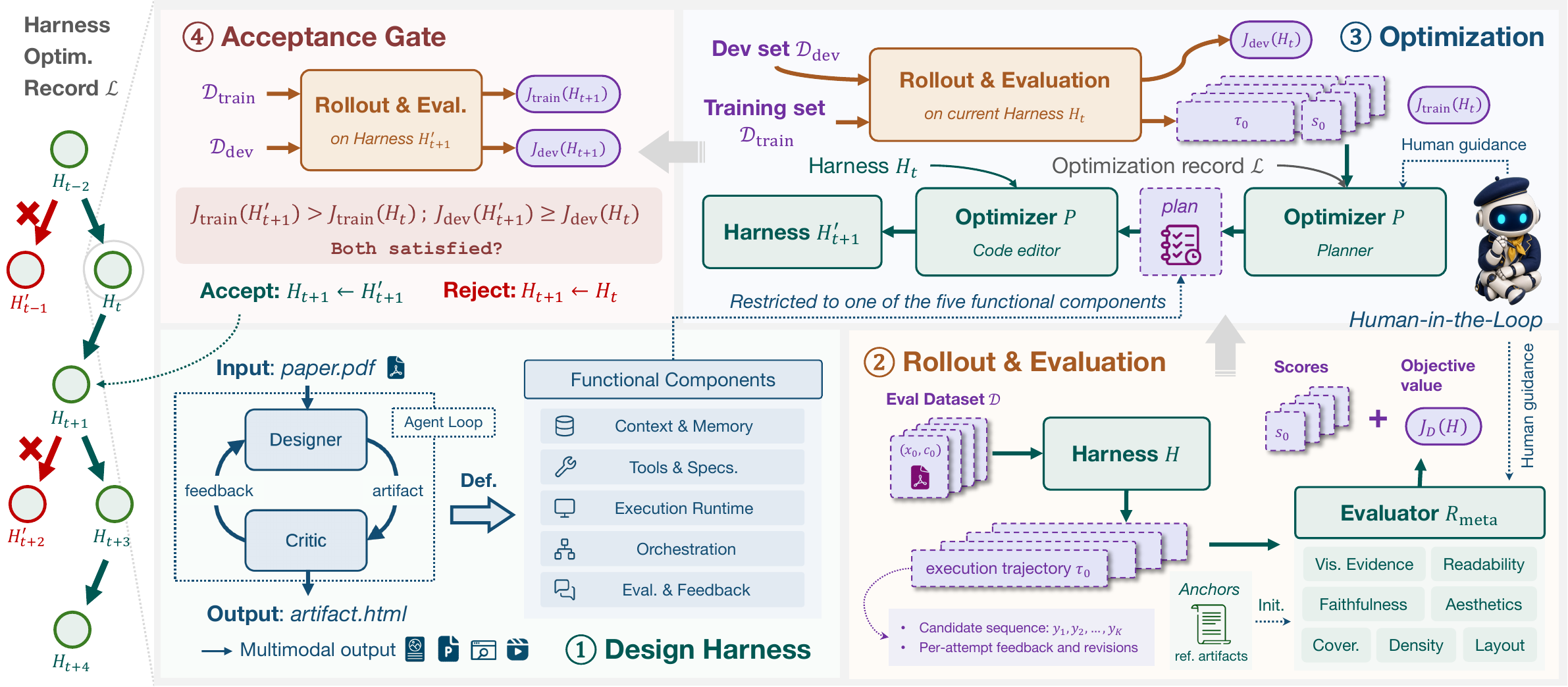}
  \caption{\textbf{Overview of AutoDesign}. The design harness iteratively generates and revises an artifact using critic feedback, thereby constituting the inner loop. The outer loop improves the design harness by running it on design tasks, evaluating its outputs, proposing an update to one component, and accepting or rejecting the candidate update. This process can run autonomously, with optional human guidance.
  }
  \label{fig:meta-harness}
\end{figure}

To close this gap, we propose \texttt{AutoDesign}, which frames human-aligned design generation as a meta-harness optimization problem \citep{robeyns2025selfimproving,lee2026metaharness,lee2026rhi,zhang2026selfharness,lin2026ahe,ren2026selfimprovement}: an agentic system recursively optimizes the design harness itself, rather than an individual artifact, based on an evaluation grounded in human preferences, \eg, annotated reference artifacts or natural language guidance. As illustrated in \Cref{fig:meta-harness}, \texttt{AutoDesign} operates through two nested loops. The inner loop is the design harness, which transforms the source context into an editable output and iteratively revises it under critic feedback. The outer loop is the meta-harness, which optimizes the design harness across tasks. It first grounds human preferences by initializing an evaluator from annotated reference artifacts. Given this human-aligned evaluator, the meta-harness aggregates rollouts and evaluation scores across tasks to identify recurrent failures, and directs a coding agent as the optimizer to propose a bounded update to the current design harness at each iteration. To prevent overfitting, the update is admitted through an acceptance gate, which accepts it only when it improves performance on the training set without degrading performance on the development set \citep{nguyen2026rsea}.

Accepted updates accumulate to \texttt{DesignHarness}, an executable system that autonomously ingests the source, generates and iteratively refines the artifact under critic feedback, and finalizes it
into a readable, visually coherent artifact. We instantiate this system for academic paper-to-poster generation, a challenging design task that must condense long, multimodal scientific sources into a single legible, visually coherent poster while preserving traceable evidence \citep{jaisankar2025poster,pang2025paper2poster,sun2025p2p,vinaykumar2026any2poster}. In this task, \texttt{DesignHarness} can produce ready-to-use conference posters that align closely with human preferences while reducing the time and manual effort required for high-quality poster production.

Evaluating a design harness requires a benchmark and evaluation protocol that jointly measure source fidelity, dense scientific communication, and rendered usability. Existing paper-to-poster benchmarks each cover only part of these aspects, such as layout, extraction, faithfulness, or visual quality, and fall short of a comprehensive, task-level protocol \citep{wang2024scipostlayout,jaisankar2025poster,pang2025paper2poster,sun2025p2p,vinaykumar2026any2poster}. \texttt{PosterBench} addresses this gap with a 100-paper spanning five disciplines, together with \texttt{PosterBench-mini}, a 10-paper subset for rapid testing. Its seven-dimensional rubric combines rule-based algorithm checks, rubric VLM judgments, and hybrid methods where both signals apply. One main track evaluates the complete system, while three controlled tracks isolate component-level effects. A system-blind human study independently validates the automatic protocol.

Across \texttt{PosterBench-mini} and \texttt{PosterBench} evaluations, the learned \texttt{DesignHarness} improves system quality while remaining practical to deploy. On \texttt{PosterBench-mini}, attaching \texttt{DesignHarness} to each of seven Code Agents raises the average \texttt{PosterBench} Score from 54.99 to 67.39 (+12.40 points). On the \texttt{PosterBench} Main Track, \texttt{AutoDesign} scores 78.32; under the same Claude Code and Claude~4.8 configuration, it outperforms the closed-source commercial system Claude Design by 7.45 points. \texttt{AutoDesign} also makes strong paper-to-poster generation accessible at low cost: with \texttt{DesignHarness}, LongCat-2.0 reaches 55.13 at approximately \$0.27 per poster\footnote{The reported cost reflects LongCat-2.0's pricing policy at evaluation time: cached context incurs no charge on a cache hit.}. Finally, across 933 valid system-blind pairwise judgments, \texttt{AutoDesign} receives the highest Bradley--Terry preference estimate, 64.0\% (95\% interval: 55.2--77.8\%); for pairs separated by at least 20 \texttt{PosterBench} points, participants prefer the \texttt{PosterBench}-preferred poster in 74.4\% of cases.

Our contributions are:

\begin{enumerate}[leftmargin=*,noitemsep]
  \item \textbf{AutoDesign}, a meta-harness optimization framework
    that turns a static design harness into a recursively improving system for
    human-aligned multimodal design. Through 7 days of evolving traces, it invokes 224 subagents, records at least 123 recursive iterations, accumulating 54 harness updates that recursively convert human-designed reference artifacts,
    rollout traces, rendering diagnostics, and evaluator feedback into persistent
    design priors for a design harness. 
  \item \textbf{DesignHarness},  the executable academic paper-to-poster system
  evolved by \texttt{AutoDesign}. It grounds a tool-using \texttt{Designer} in
  paper context and combines editable generation with rendering,
  rule-based validation, and visual critic feedback for localized revision,
  producing source-grounded posters that remain directly usable and editable.
  \item \textbf{PosterBench}, a comprehensive evaluation protocol for paper-to-poster evaluation. It evaluates scientific communication quality and
  executable-artifact reliability through a seven-dimensional rubric spanning
  faithfulness, coverage, density, visual evidence, layout, readability, and
  aesthetics. On \texttt{PosterBench}, the state-of-the-art coding agent Claude Code (Claude 4.8) achieves 70.01; attaching DesignHarness can improve it by +8.31, also surpassing the best commercial design agent Claude Design by 7.45 points.
  \item Under a fully autonomous long-horizon agentic loop, \texttt{DesignHarness} produces human-level academic posters. In a poster generation run, it executes 253 tool calls and 11 editing turns within 40 minutes for less than \$3, with negligible human intervention. \href{https://designanything.ai/}{Demo} is it available as a research-preview platform for interactive use and localized
  revision of editable posters.
\end{enumerate}

\section{Meta-Harness Formulation}
\label{sec:task}

\subsection{Design Harness Definition}
\label{sec:design-harness-definition}

Following recent work that treats harnesses as an optimization target distinct from model weights \citep{lee2026metaharness,ren2026selfimprovement}, we define a \emph{design harness} $H$ as the system surrounding a fixed model that turns a multimodal source (\eg, an academic paper or report) into a human-facing artifact (\eg, a presentation slide, poster, video, or web page):
\begin{equation}
  y \sim H(\pi_\theta,x,c),
  \label{eq:harness-rollout}
\end{equation}
where $\pi_\theta$ is an LLM or MLLM, $x$ is the multimodal input, and $c$ specifies the context including the target medium and user constraints. The harness produces an artifact $y$ through an execution trajectory $\tau$, which records the sequence of intermediate actions, states, and revisions leading to the final output.

To enable systematic meta-harness optimization and facilitate credit assignment, we decompose the design harness $H$ into five functional components:
\begin{itemize}[leftmargin=*]
  
  \item \textbf{Context and Memory}: source management, prompts, skills, reusable assets and persistent state.
  \item \textbf{Tools and Specifications}: tools and editable artifact specifications for layout, typography, and provenance.
  \item \textbf{Execution Runtime}: the workspace and runtime for authoring, rendering, validating, and exporting artifacts.
  \item \textbf{Orchestration}: task routing, attempt budgets, loop control, candidate selection, fallback, and finalization.
  \item \textbf{Evaluation and Feedback}: rule-based validation, model-based critique, and localized feedback for revision.
\end{itemize}
This decomposition specifies only the high-level abstraction of the design harness. The concrete implementation of each component is instantiated and iteratively improved by the meta-harness.

\subsection{Meta-Harness Definition}
\label{sec:meta-harness-definition}

We define a \emph{meta-harness} as a system that operates on the design harness \citep{lee2026metaharness,zhang2026selfharness,lin2026ahe}. Given a user specification $q$ and an optional initial design harness $H_0$, the meta-harness instantiates or iteratively improves the concrete implementation of the design harness, yielding an optimized harness $H_T$. Unlike a design harness, which transforms multimodal sources into human-facing artifacts, the meta-harness
transforms harness requirements and implementations into an improved design harness.

The optimization target is the expected quality of the artifacts produced by the design harness. Let $p_{\mathrm{task}}$ denote the task distribution. We define the performance of a design harness $H$ as
\begin{equation}
  J(H)
  =
  \mathbb{E}_{(x,c)\sim p_{\mathrm{task}},\,
  y\sim H(\pi_\theta,x,c)}
  \left[R_{\mathrm{meta}}(y,x,c)\right],
  \label{eq:harness-score}
\end{equation}
where $R_{\mathrm{meta}}(y,x,c)$ denotes the evaluator used by the meta-harness to assess the quality of artifact $y$ with respect to its source $x$ and design context $c$. Accordingly, the meta-harness optimization objective is
\begin{equation}
  H^\star = \underset{H}{\operatorname{arg\,max}}\;J(H).
  \label{eq:harness-objective}
\end{equation}
Throughout this process, the parameters $\theta$ of the underlying model $\pi_\theta$ in the harness remain fixed. The optimization therefore acts on the system surrounding the model rather than on the model itself, consistent with the model-versus-scaffold distinction in recent self-improving-agent taxonomies \citep{ren2026selfimprovement}.

\section{Meta-Harness Learning Loop}
\label{sec:meta}

\texttt{AutoDesign} organizes artifact generation (\ie, design harness) and harness optimization (\ie, meta-harness) as two nested feedback loops, as illustrated in \Cref{fig:meta-harness-loop}. For each design task $(x,c)$, the design harness executes an \emph{inner loop} that repeatedly generates and revises the current artifact. This loop operates on the artifact $y$ under a fixed design harness $H$ and records the resulting execution trajectory $\tau$. Across tasks, the meta-harness executes an \emph{outer loop} that analyzes these trajectories and their evaluation results to identify recurrent failures and update the design harness $H$. Thus, the inner loop improves a single artifact without changing $H$, whereas the outer loop improves $H$ based on evidence collected from multiple generation runs.

\subsection{Inner Loop}
\label{sec:inner-loop}

\begin{wrapfigure}[20]{r}{0.4\columnwidth}
  \centering
  \includegraphics[width=\linewidth]{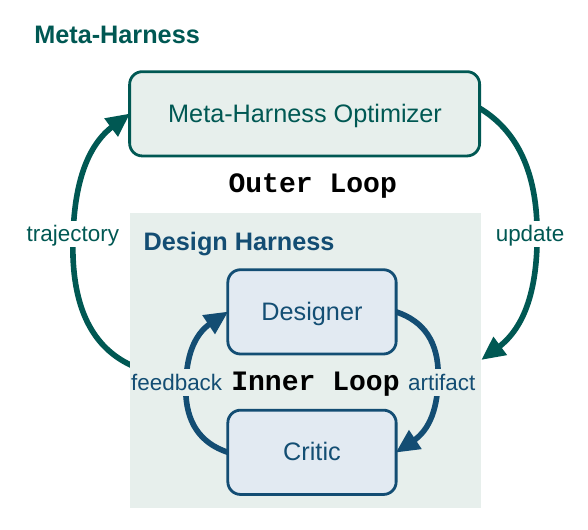}
  \caption{The inner and outer loops of the \texttt{AutoDesign} Framework. The inner loop updates the artifact according to the design harness, while the outer loop updates the design harness.}
  \label{fig:meta-harness-loop}
\end{wrapfigure}

We initialize the design harness with a minimal inner-loop scaffold consisting of two abstract modules: a \emph{designer} $M_{\mathrm{design}}$ and a \emph{critic} $M_{\mathrm{critic}}$. As shown in \Cref{fig:meta-harness-loop}, the designer generates and revises the artifact, while the critic evaluates its current state and provides feedback for the next revision. At refinement step $k$, their interaction is defined as
\begin{equation}
  \begin{aligned}
    y_k &= M_{\mathrm{design}}(y_{k-1},f_{k-1};x,c), \\
    f_k &= M_{\mathrm{critic}}(y_k;x,c),
  \end{aligned}
  \label{eq:inner-loop}
\end{equation}
where $y_k$ is the artifact at step $k$ and $f_k$ is the corresponding feedback, with $y_0$ and $f_0$ left empty so that the first step produces an initial draft from $(x,c)$ alone. Repeated evaluation and revision produce an execution trajectory $\tau$ under the current design harness.

This formulation specifies only the roles of the two modules and the information flow between them. It defines the basic structure of the initial design harness while leaving its concrete realization open. During outer-loop optimization, the meta-harness may refine the two modules and their interaction, including their prompts, tools, feedback mechanisms, and loop-control policies, based on observed task performance.

\subsection{Outer Loop}
\label{sec:outer-loop}

The outer loop improves the design harness across tasks, as illustrated in \Cref{fig:meta-harness}. Each iteration proceeds in four stages: rollout, evaluation, update proposal, and acceptance.

\noindent\textbf{Rollout.} At outer-loop iteration $t$, the current design harness $H_t$ is executed on a training task set $\mathcal{D}_{\mathrm{train}}=\{(x_i,c_i)\}_{i=1}^{N_{\mathrm{train}}}$, where $x_i$ is a multimodal source and $c_i$ specifies the target medium and the corresponding design requirements. Each execution produces an artifact $y_t^i$ and a corresponding trajectory $\tau_t^i$. We denote the collection of trajectories by $\boldsymbol{\tau}_t=\{\tau_t^i\}_{i=1}^{N_{\mathrm{train}}}$.

\noindent\textbf{Evaluation.} Before outer-loop optimization, we provide an evaluator coding agent with reference artifacts annotated by humans along seven quality dimensions: (i) Faithfulness, (ii) Coverage, (iii) Density, (iv) Visual Evidence, (v) Layout, (vi) Readability, and (vii) Aesthetics. The agent uses these examples to implement the evaluator $R_{\rm meta}$, combining rule-based checks for directly measurable properties with VLM-based judgments for perceptual properties such as aesthetics. Once constructed, $R_{\rm meta}$ remains fixed during autonomous optimization. The resulting evaluator is then used to assess every generated artifact, $s_t^i=R_{\mathrm{meta}}(y_t^i,x_i,c_i)$, and the resulting scores are collected into a batch denoted by $\boldsymbol{s}_t$.
This optimization-time evaluator is distinct from the frozen \texttt{PosterBench} protocol used for final system comparison; the latter is specified in \Cref{app:poster-evaluation-criteria}.

\noindent\textbf{Update proposal.} We denote the \emph{meta-harness optimizer} by $P$. In addition to the current design harness and the evidence collected at the current iteration, the meta-harness maintains an optimization record $\mathcal{L}$, serving as persistent context across outer-loop iterations. At iteration $t$, the optimizer takes the current harness $H_t$, the collected trajectories $\boldsymbol{\tau}_t$, their evaluation scores $\boldsymbol{s}_t$, and the optimization record $\mathcal{L}$ as input, and produces a
candidate updated harness $H'_{t+1}$:
\begin{equation}
  H'_{t+1}
  =
  P\left(
    H_t,
    \boldsymbol{\tau}_t,
    \boldsymbol{s}_t,
    \mathcal{L}
  \right).
  \label{eq:outer-loop}
\end{equation}
The prime indicates that $H'_{t+1}$ is an update proposal, whose acceptance is determined by the subsequent gating stage.

\begin{figure}[t]
  \centering
  \includegraphics[width=\textwidth]{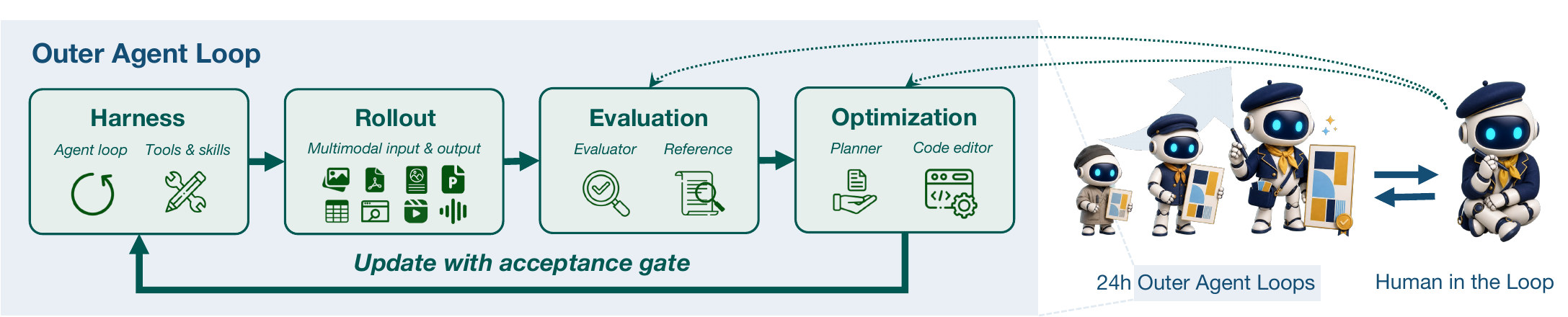}
  \caption{\textbf{AutoDesign outer loop.} Each iteration proceeds through rollout, evaluation, update proposal, and acceptance. The loop runs autonomously, with optional human guidance to redirect the evaluation and optimization.}
  \label{fig:autodesign-method-detail}
\end{figure}

We instantiate $P$ as a coding agent that sequentially assumes the roles of a planner and a code editor, as illustrated in \Cref{fig:meta-harness}. In the planner role, the agent analyzes the current trajectories and scores together with the optimization history in $\mathcal{L}$. It dispatches parallel subagents to inspect the trajectories and their scores, synthesizes their findings into structured evidence of recurrent failures, and formulates a harness update plan. The plan specifies the observed failure modes, the harness component to be modified, and the intended changes. In the code-editor role, the agent implements these changes in the current design harness $H_t$, yielding the candidate harness $H'_{t+1}$.

Each outer-loop iteration is restricted to exactly one of the five harness components defined in \Cref{sec:design-harness-definition}. An update may span multiple files within the selected component, but it cannot modify another component in the same iteration. This restriction keeps credit assignment interpretable, as each gain or regression is attributable to a single coherent intervention rather than to several simultaneous changes.

\noindent\textbf{Acceptance gate.} Once an update has been proposed, the meta-harness determines whether it should replace the current harness through a separate acceptance gate. Let $J_{\mathrm{train}}$ and $J_{\mathrm{dev}}$ denote the objective in \Cref{eq:harness-score} evaluated on the training set $\mathcal{D}_{\mathrm{train}}$ and on an independent development set $\mathcal{D}_{\mathrm{dev}}$, respectively. A candidate is accepted only when its performance on $\mathcal{D}_{\mathrm{train}}$ improves and its performance on $\mathcal{D}_{\mathrm{dev}}$ does not decline:
\begin{equation}
  \operatorname{Accept}(H'_{t+1})
  \iff
  J_{\mathrm{train}}(H'_{t+1})>J_{\mathrm{train}}(H_t)
  \ \land\
  J_{\mathrm{dev}}(H'_{t+1})\ge J_{\mathrm{dev}}(H_t).
  \label{eq:acceptance}
\end{equation}
If the condition holds, the meta-harness promotes $H'_{t+1}$ to $H_{t+1}$; otherwise $H_t$ is retained. Results on the development set are used exclusively by the acceptance gate and are never exposed to $P$ when constructing an update proposal. Therefore, $\mathcal{D}_{\mathrm{dev}}$ serves as a guard against overfitting the harness to the training tasks.

After the acceptance decision, the meta-harness appends the completed iteration to the optimization record $\mathcal{L}$. For each iteration $t$, $\mathcal{L}$ stores the harness $H_t$, the trajectories and scores, the selected harness component, the update plan and the corresponding code changes, and the acceptance decision, with a repository checkpoint preserving the harness implementation at that iteration. Trajectories and scores from the development set are not included in the record.

The updated $\mathcal{L}$ is supplied to $P$ as persistent context in the next iteration. When a candidate is rejected, the record allows the next iteration to propose a different update while retaining the evidence of what has already been tried. $\mathcal{L}$ thus supports comparison, reproducibility, and rollback across iterations. Note that the outer loop maintains a single active harness at each iteration and does not perform tree search over harness variants. The complete meta-harness optimization procedure is summarized in \Cref{alg:meta-harness}.

\begin{algorithm}[t]
\caption{\texttt{AutoDesign} meta-harness optimization}
\label{alg:meta-harness}
\begin{algorithmic}[1]
\REQUIRE Fixed model $\pi_\theta$; initial design harness $H_0$; evaluator $R_{\rm meta}$
\REQUIRE Training task set $\mathcal{D}_{\mathrm{train}}$; development task set $\mathcal{D}_{\mathrm{dev}}$; outer-loop iterations $T$
\ENSURE Optimized design harness $H_T$ and optimization record $\mathcal{L}$
\STATE Run $H_0$ on $\mathcal{D}_{\mathrm{train}}$; collect execution trajectories $\boldsymbol{\tau}_0$ and evaluator scores $\boldsymbol{s}_0$
\STATE Run $H_0$ on $\mathcal{D}_{\mathrm{dev}}$ and collect evaluator scores $\boldsymbol{s}^{\mathrm{dev}}_{0}$
\FOR{$t=0$ to $T-1$}
  \STATE The meta-harness optimizer $P$ inspects $\boldsymbol{\tau}_t$, $\boldsymbol{s}_t$, and $\mathcal{L}$ and proposes a candidate updated harness $H'_{t+1}$
  \STATE Run $H'_{t+1}$ on $\mathcal{D}_{\mathrm{train}}$; collect execution trajectories $\boldsymbol{\tau}'_{t+1}$ and evaluator scores $\boldsymbol{s}'_{t+1}$
  \STATE Run $H'_{t+1}$ on $\mathcal{D}_{\mathrm{dev}}$ and collect evaluator scores $\boldsymbol{s}^{\prime\mathrm{dev}}_{t+1}$
  \IF{$J_{\mathrm{train}}(H'_{t+1})>J_{\mathrm{train}}(H_t)$ \AND $J_{\mathrm{dev}}(H'_{t+1})\ge J_{\mathrm{dev}}(H_t)$}
    \STATE $d_t\leftarrow\textsc{Accept}$
    \STATE $(H_{t+1}, \boldsymbol{\tau}_{t+1}, \boldsymbol{s}_{t+1}, \boldsymbol{s}^{\rm dev}_{t+1}) \leftarrow (H'_{t+1}, \boldsymbol{\tau}'_{t+1}, \boldsymbol{s}'_{t+1}, \boldsymbol{s}'^{\rm dev}_{t+1})$
  \ELSE
    \STATE $d_t\leftarrow\textsc{Reject}$
    \STATE $(H_{t+1}, \boldsymbol{\tau}_{t+1}, \boldsymbol{s}_{t+1}, \boldsymbol{s}^{\rm dev}_{t+1}) \leftarrow (H_t, \boldsymbol{\tau}_t, \boldsymbol{s}_t, \boldsymbol{s}^{\rm dev}_t)$
  \ENDIF
  \STATE Append the harness checkpoint and iteration record to $\mathcal{L}$
\ENDFOR
\STATE \textbf{return} $(H_T,\mathcal{L})$
\end{algorithmic}
\end{algorithm}

\noindent \textbf{Human-in-the-Loop.} \texttt{AutoDesign} also supports an optional human-in-the-loop mode, as illustrated in \Cref{fig:meta-harness}. Human intervention can operate through two channels. First, at iteration $t$, a user may provide directional guidance $g_t$ in natural language, which is supplied to the planner alongside the trajectories and evaluation scores, so that the update proposal becomes $H'_{t+1}=P(H_t,\boldsymbol{\tau}_t,\boldsymbol{s}_t,\mathcal{L},g_t)$. We introduce this mechanism because the coding agent acting as $P$ may converge prematurely to a locally satisfactory harness configuration, at which point outer-loop optimization stagnates. Guidance can inject task-specific heuristics or redirect the search toward alternative improvements.

Second, human guidance may be provided to the coding agent responsible for implementing the evaluator when visual inspection reveals a systematic artifact bias not captured by $R_{\rm meta}$. Such evaluator revision requires explicit human input. Otherwise, $R_{\rm meta}$ remains fixed because the meta-harness receives no external signal with which to identify or correct evaluator bias.

In both cases, the human provides observations or high-level directions rather than directly editing the harness or evaluator implementation. When no guidance is provided, the outer loop operates autonomously as defined in \Cref{eq:outer-loop}.

\FloatBarrier

\section{The Optimized DesignHarness}
\label{sec:harness}

\begin{figure}[t]
  \centering
  \includegraphics[width=\textwidth]{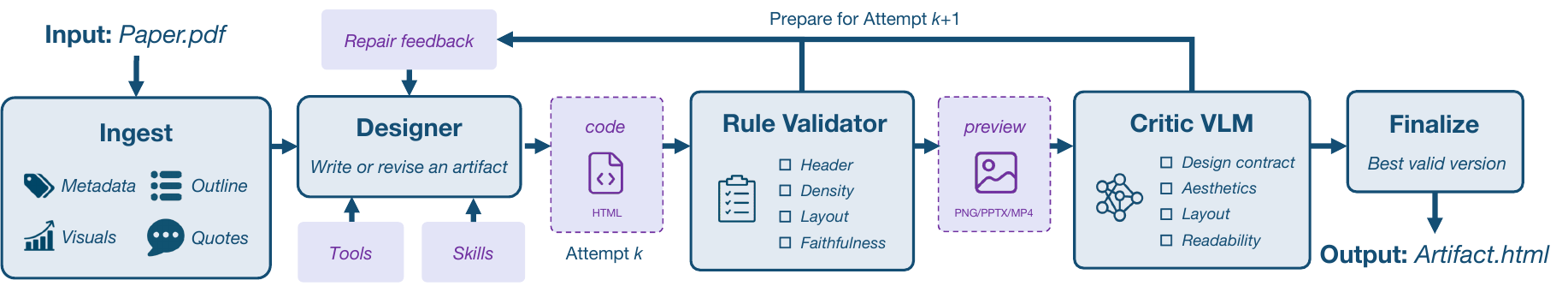}

  \caption{\textbf{Overview of DesignHarness}, optimized for human-facing artifact generation. Given source materials, the designer iteratively generates and revises an editable artifact using feedback from dual critics: a rule-based validator and a model-based visual critic. \texttt{DesignHarness} then finalizes the best valid candidate.}
  \label{fig:autodesign-method}
\end{figure}

The meta-harness optimization described in \Cref{sec:meta} yields \texttt{DesignHarness}. The resulting implementation supports multiple output media, including academic posters, presentation slides, videos, and web pages. We characterize its resulting architecture by examining the final implementation obtained through meta-harness optimization. As summarized in \Cref{fig:autodesign-method}, our analysis identifies four main stages: source ingestion, iterative artifact generation and revision by a designer module, feedback from a rule-based validator and a VLM-based critic, and finalization that prepares the selected candidate for delivery.

\subsection{Paper Ingestion}
\label{sec:source-ingestion}

The ingestion stage transforms the input source $x$ and the design context $c$ into a structured, provenance-aware context for subsequent generation and revision. It extracts the document metadata and section outline, identifies key passages supporting the main claims, and records figures and tables together with their source locations, reflecting the joint textual and visual evidence selection required in paper-to-poster generation \citep{jaisankar2025poster,pang2025paper2poster,sun2025p2p}. These materials are then organized into a content brief and a medium-specific artifact plan, which specify the target output format, the claims to be conveyed, and the visual evidence supporting each of them. Every extracted element retains a reference to its location in $x$, so that source-derived statements and visual materials used in the artifact can be traced back to the source and checked during revision. The resulting context is constructed once, retained across all inner-loop refinement steps, and provided to the designer as source-grounded input.

\subsection{Artifact Generation and Revision}
\label{sec:artifact-generation}

The designer module is implemented as a coding agent that generates or revises the artifact from the ingested source context using the tools and skills available in the harness. At refinement step $k$, it conditions on the current artifact $y_{k-1}$, the feedback $f_{k-1}$, and the ingested context to produce the next candidate $y_k$, thereby instantiating $M_{\mathrm{design}}$ in \Cref{eq:inner-loop}. Consistent with layer- and code-based design generation, the artifact remains as editable HTML files throughout refinement \citep{qu2025igd,liu2026designascode}, allowing revisions to be implemented as localized code edits without requiring regeneration of the entire output. For visual critique, it can be rendered or exported as a medium-specific preview, such as PNG, PPTX, or MP4.

\subsection{Validation and Finalization}
\label{sec:validation-revision}

At each refinement step $k$, the candidate artifact $y_k$ produced by the designer is examined by the rule-based validator, which applies a set of deterministic blocking checks. At a high level, these checks determine whether the candidate satisfies the requirements for terminating refinement and cover issues such as unsafe or missing assets, broken provenance links between incorporated materials and their sources, severe overflow or overlap, and violations of the required typographic and layout constraints. If the candidate passes all of them, the inner loop terminates and the candidate proceeds directly to finalization. Otherwise, the validator returns localized diagnostics for the detected violations, together with the results of non-blocking checks on properties such as content coverage, information density, and numerical consistency with the source.

When a candidate fails the blocking checks, it is also rendered into a medium-specific preview and inspected by a critic VLM. The critic assesses rendered properties of the candidate, including compliance with the design context, layout, readability, and aesthetics. The two sources of feedback are consolidated into the repair signal $f_k$ and passed to the designer for the next attempt. This feedback-to-revision pattern is related to recursive self-refinement and agent-as-a-judge approaches \citep{madaan2023selfrefine,zhuge2025agentasjudge}. In the notation of \Cref{eq:inner-loop}, rule-based validation and visual critique jointly instantiate $M_{\mathrm{critic}}$.

The final implementation permits at most $K=12$ refinement attempts. As soon as a candidate $y_k$ passes all blocking checks, the attempt loop terminates and the candidate is passed to the finalization stage. This stage applies the remaining post-processing, such as final rendering adjustments, mathematical typesetting, and inlining of referenced assets, to produce a self-contained output. If the attempt budget is exhausted without any candidate passing all blocking checks, the harness uses the retained attempt history and applies a sequence of fallback mechanisms to identify a deliverable candidate while retaining essential safety and integrity constraints. The selected candidate is then passed to the same finalization stage. The sequence of candidate generation, validation, critique, and revision constitutes the execution trajectory $\tau$ in \Cref{sec:inner-loop}, while the design harness $H$ remains fixed throughout this process.

\FloatBarrier

\section{Evaluation Protocol}
\label{sec:evaluation}

\subsection{Benchmark: PosterBench}
\label{sec:benchmark}

\begin{figure*}[t]
  \centering
  \includegraphics[width=0.98\textwidth]{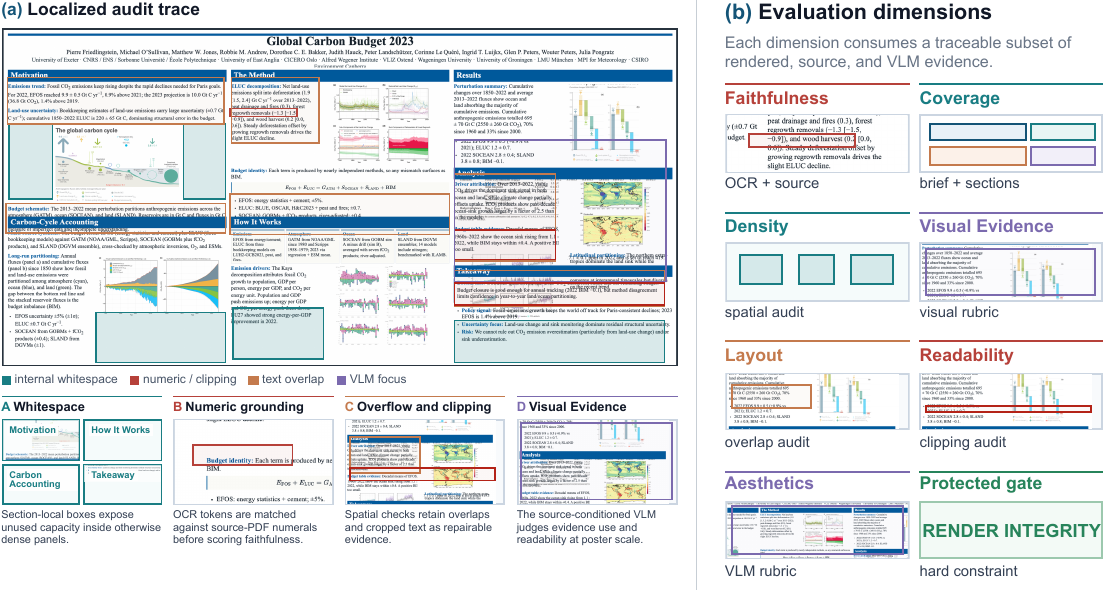}
  \caption{\textbf{PosterBench evaluation protocol.} \texttt{PosterBench} combines
    rule-based algorithms for traceable spatial, OCR, numeric-grounding, and
    render-integrity checks with rubric-guided VLM judges for source-grounded
    assessment of visual evidence, layout, readability, and aesthetics.}
  \label{fig:evaluation-protocol-trace}
\end{figure*}

\texttt{PosterBench} contains a 100-paper Main Track and \texttt{PosterBench-mini}, a shared
10-paper subset reported in \Cref{tab:small-main-results}. The papers span five disciplines: \textit{AI/ML}, \textit{biomedicine and health},
\textit{climate and earth environment}, \textit{economics and policy}, and \textit{physics and
astronomy}. Every system receives the same source paper and associated source
assets, and its output is rendered to a common poster format before scoring.
The shared paper-to-poster generation interface and the exact fixed-versus-varied
factors for every reported track are documented in
\Cref{app:shared-generation-prompt} and \Cref{tab:appendix-tracks}.

For paper $p_i$ and candidate artifact $A_i$, the fixed \texttt{PosterBench} evaluator
returns $\mathbf q_i\in[0,10]^7$, ordered as Faithfulness, Coverage, Density,
Visual Evidence, Layout, Readability, and Aesthetics. It forms the weighted
rubric score
\begin{equation}
R_{\mathrm{rubric}}(p_i,A_i)
= \sum_{j=1}^{7}\alpha_j q_{i,j}/10,
\qquad
\boldsymbol{\alpha}=(10,10,15,10,20,25,10).
\label{eq:posterbench-rubric}
\end{equation}
The final score applies the strictest active record-level ceiling before
benchmark averaging:
\begin{equation}
\begin{aligned}
R_{\mathrm{poster}}(p_i,A_i)
&= \min\!\Bigl(R_{\mathrm{rubric}}(p_i,A_i), C_i^{\mathrm{layout}},\\
&\hspace{3.3em} C_i^{\mathrm{viability}}, C_i^{\mathrm{failure}},
C_i^{\mathrm{gate}}\Bigr),\\
\mathrm{Overall}
&=\frac{1}{N}\sum_{i=1}^{N}R_{\mathrm{poster}}(p_i,A_i).
\end{aligned}
\label{eq:posterbench-aggregation}
\end{equation}
Here, $C_i^{\mathrm{layout}}$, $C_i^{\mathrm{viability}}$,
$C_i^{\mathrm{failure}}$, and $C_i^{\mathrm{gate}}$ bound severe layout
damage, insufficient presentation viability, confirmed visible failures, and
protected render-integrity violations, respectively; inactive ceilings are
100. A standard P0 gate caps a score at 40, and more severe gate types may set a
lower cap. Thus, the metric columns are dimension means, while \textbf{Overall}
is the mean of capped poster scores and cannot generally be recovered by
reweighting those displayed means.

\texttt{PosterBench} is a frozen external evaluator, separate from the optimization-time
evaluator $R_{\mathrm{meta}}$ used in the meta-harness outer loop
(\Cref{sec:outer-loop}). $R_{\mathrm{meta}}$ supplies feedback for updating
\texttt{DesignHarness}; \texttt{PosterBench} evaluates completed systems and is neither
optimized nor modified by the outer loop. The complete operational rubric,
aggregation, protected gates, and per-case record schema are specified in
\Cref{app:poster-evaluation-criteria}.

\Cref{fig:evaluation-protocol-trace} shows the evaluator's evidence path
on a rendered poster. Programmatic spatial and OCR audits localize unused
regions, text overflow or clipping, and source-inconsistent numeric claims.
The source-conditioned VLM then assesses paper-grounded visual quality from the
rendered artifact and paper brief. These
dimension-level signals are aggregated with fixed weights and record-level
ceilings.

\noindent\textbf{PosterBench Main Track.}
\Cref{tab:main-results} reports the 100-paper \texttt{PosterBench} Main Track, comparing design
agents, standalone coding agents, and task-specific handcrafted workflow under the fixed
\texttt{PosterBench} evaluator. \texttt{AutoDesign} attains the highest \texttt{PosterBench} Score at
78.32. Under the matched Claude Code and Claude~4.8 configuration, \texttt{AutoDesign}
scores 78.32, exceeding Claude Design by 7.45 points and OpenDesign by 8.87
points.
Curated qualitative comparisons, which are excluded from the aggregate
\texttt{PosterBench} results, appear in \Cref{fig:appendix-qualitative-matrix}.

\begin{table}[t]
  \centering
  \caption{\texttt{PosterBench} Score and dimension scores for the \texttt{PosterBench} Main
  Track on the 100-paper evaluation set. Systems are grouped by their
  primary design mechanism. ``---'' denotes a layer that a system does not use.}
  \scriptsize
  \setlength{\tabcolsep}{2pt}
  \renewcommand{\arraystretch}{1.12}
  \begin{tabularx}{\textwidth}{@{}
    >{\raggedright\arraybackslash}p{0.115\textwidth}
    >{\raggedright\arraybackslash}p{0.115\textwidth}
    >{\raggedright\arraybackslash}p{0.12\textwidth}
    >{\raggedright\arraybackslash}p{0.148\textwidth}
    *{8}{>{\centering\arraybackslash}X}@{}}
    \toprule
    \textbf{System} & \mbox{\textbf{Design Harness}} & \textbf{Coding Agent} & \textbf{Model} &
    \textbf{Score} & \textbf{Faith.} & \textbf{Cover.} & \textbf{Density} &
    \textbf{Vis. Ev.} & \textbf{Layout} & \textbf{Read.} & \textbf{Aesth.} \\
    \midrule
    \multicolumn{12}{@{}l}{\textit{\textbf{Design Agent}}} \\
    \addlinespace[2pt]
    \rowcolor[HTML]{E9EEFF}
    \textbf{AutoDesign} & \mbox{\textbf{DesignHarness}} & Claude Code & Claude 4.8 & 78.32 & 9.35 & 9.40 & 8.41 & 5.97 & 8.55 & 8.17 & 5.59 \\
    \rowcolor[HTML]{E9EEFF}
    \textbf{AutoDesign} & \mbox{\textbf{DesignHarness}} & Codex & GPT 5.5 & 77.97 & 9.57 & 9.35 & 8.63 & 5.47 & 7.75 & 8.08 & 7.16 \\
    \mbox{Claude Design} & \mbox{Claude Design} & \mbox{Claude Code} & \mbox{Claude 4.8} & 70.87 & 9.22 & 9.90 & 6.48 & 7.62 & 8.08 & 5.96 & 7.36 \\
    OpenDesign & OpenDesign & Claude Code & Claude 4.8 & 69.45 & 9.17 & 9.37 & 7.51 & 6.12 & 7.08 & 6.19 & 7.01 \\
    OpenDesign & OpenDesign & Codex & GPT 5.5 & 62.17 & 9.24 & 8.49 & 6.91 & 5.05 & 7.24 & 5.76 & 6.00 \\
    \midrule
    \multicolumn{12}{@{}l}{\textit{\textbf{Coding Agent}}} \\
    \addlinespace[2pt]
    Codex & --- & Codex & GPT 5.5 & 73.37 & 9.64 & 9.71 & 7.69 & 7.43 & 8.03 & 6.08 & 6.60 \\
    \mbox{Claude Code} & --- & \mbox{Claude Code} & \mbox{Claude 4.8} & 70.01 & 9.25 & 9.88 & 5.71 & 7.01 & 9.46 & 6.68 & 6.53 \\
    Doubao & --- & Claude Code & Seed 2.1 & 61.14 & 8.93 & 8.96 & 5.95 & 5.97 & 6.51 & 5.63 & 4.61 \\
    GLM & --- & \mbox{Claude Code} & GLM 5.2 & 52.22 & 8.60 & 7.44 & 5.23 & 5.29 & 7.03 & 4.90 & 5.15 \\
    Kimi & --- & \mbox{Claude Code} & \mbox{Kimi K2.7} & 51.46 & 8.33 & 7.15 & 4.72 & 5.62 & 7.26 & 4.93 & 4.81 \\
    DeepSeek & --- & \mbox{Claude Code} & \mbox{DeepSeekV4-Pro} & 46.01 & 8.23 & 6.68 & 4.10 & 3.82 & 6.95 & 4.41 & 3.38 \\
    \midrule
    \multicolumn{12}{@{}l}{\textit{\textbf{Human-Crafted Workflow}}} \\
    \addlinespace[2pt]
    PosterGen & --- & --- & Claude 4.8 & 56.71 & 8.84 & 8.25 & 4.31 & 5.62 & 8.61 & 5.36 & 5.28 \\
    Any2Poster & --- & --- & Claude 4.8 & 49.09 & 8.26 & 5.44 & 4.18 & 3.65 & 9.59 & 4.47 & 3.10 \\
    \mbox{Paper2Poster} & --- & --- & \mbox{Claude 4.8} & 44.61 & 6.35 & 2.35 & 8.36 & 2.16 & 3.69 & 4.87 & 1.69 \\
    \bottomrule
  \end{tabularx}

  \label{tab:main-results}
\end{table}

\noindent\textbf{PosterBench-mini Main Track.}
\Cref{tab:small-main-results} reports results on \texttt{PosterBench-mini}, the original
10-paper subset used for the controlled ablations.

\begin{table}[t]
  \centering
  \caption{\texttt{PosterBench} Score and dimension scores for the \texttt{PosterBench-mini} Main
  Track on the original 10-paper subset.}
  \scriptsize
  \setlength{\tabcolsep}{2pt}
  \renewcommand{\arraystretch}{1.12}
  \begin{tabularx}{\textwidth}{@{}
    >{\raggedright\arraybackslash}p{0.115\textwidth}
    >{\raggedright\arraybackslash}p{0.115\textwidth}
    >{\raggedright\arraybackslash}p{0.12\textwidth}
    >{\raggedright\arraybackslash}p{0.148\textwidth}
    *{8}{>{\centering\arraybackslash}X}@{}}
    \toprule
    \textbf{System} & \mbox{\textbf{Design Harness}} & \textbf{Coding Agent} & \textbf{Model} &
    \textbf{Score} & \textbf{Faith.} & \textbf{Cover.} & \textbf{Density} &
    \textbf{Vis. Ev.} & \textbf{Layout} & \textbf{Read.} & \textbf{Aesth.} \\
    \midrule
    \multicolumn{12}{@{}l}{\textit{\textbf{Design Agent}}} \\
    \addlinespace[2pt]
    \rowcolor[HTML]{E9EEFF}
    \textbf{AutoDesign} & \mbox{\textbf{DesignHarness}} & Codex & GPT 5.5 & \textbf{81.46} & 9.88 & 9.90 & 8.64 & 6.20 & 8.01 & 7.96 & 7.45 \\
    \rowcolor[HTML]{E9EEFF}
    \textbf{AutoDesign} & \mbox{\textbf{DesignHarness}} & Claude Code & Claude 4.8 & \textbf{74.56} & 9.28 & 8.80 & 8.15 & 5.00 & 9.00 & 7.86 & 6.75 \\
    OpenDesign & OpenDesign & Claude Code & Claude 4.8 & 70.36 & 9.10 & 9.00 & 7.39 & 6.15 & 7.13 & 6.38 & 7.31 \\
    \mbox{Claude Design} & \mbox{Claude Design} & \mbox{Claude Code} & \mbox{Claude 4.8} & 66.83 & 8.95 & 10.00 & 5.24 & 7.90 & 8.54 & 6.09 & 6.77 \\
    OpenDesign & OpenDesign & Codex & GPT 5.5 & 60.58 & 9.03 & 9.20 & 6.13 & 5.45 & 7.68 & 5.75 & 5.75 \\
    \midrule
    \multicolumn{12}{@{}l}{\textit{\textbf{Coding Agent}}} \\
    \addlinespace[2pt]
    Codex & --- & Codex & GPT 5.5 & 75.87 & 9.47 & 9.40 & 7.88 & 7.55 & 8.44 & 6.00 & 7.00 \\
    \mbox{Claude Code} & --- & \mbox{Claude Code} & \mbox{Claude 4.8} & 69.55 & 9.19 & 10.00 & 5.21 & 7.50 & 9.76 & 6.90 & 6.40 \\
    Kimi & --- & \mbox{Claude Code} & \mbox{Kimi K2.7} & 57.20 & 8.20 & 8.20 & 4.91 & 6.40 & 7.91 & 5.60 & 5.95 \\
    Doubao & --- & Claude Code & Seed 2.1 & 54.01 & 7.99 & 8.20 & 6.30 & 5.55 & 6.01 & 5.52 & 6.15 \\
    GLM & --- & \mbox{Claude Code} & GLM5.2 & 50.32 & 8.83 & 7.80 & 3.79 & 3.85 & 8.21 & 5.11 & 5.45 \\
    DeepSeek & --- & \mbox{Claude Code} & \mbox{DeepSeekV4-Pro} & 34.73 & 8.08 & 7.90 & 2.94 & 3.33 & 5.98 & 3.71 & 5.05 \\
    \midrule
    \multicolumn{12}{@{}l}{\textit{\textbf{Human-Crafted Workflow}}} \\
    \addlinespace[2pt]
    PosterGen & --- & --- & Claude 4.8 & 51.82 & 8.95 & 9.33 & 3.39 & 6.67 & 7.67 & 4.96 & 5.50 \\
    Any2Poster & --- & --- & Claude 4.8 & 46.88 & 8.09 & 4.90 & 3.69 & 3.65 & 9.55 & 4.40 & 3.20 \\
    \mbox{Paper2Poster} & ---& --- & \mbox{Claude 4.8} & 42.06 & 6.16 & 2.10 & 8.25 & 2.30 & 2.76 & 4.82 & 1.80 \\
    \bottomrule
  \end{tabularx}
  \label{tab:small-main-results}
\end{table}

On \texttt{PosterBench-mini}, \texttt{AutoDesign} reaches 81.46 with Codex,
compared with 75.87 for the native Codex baseline, and 74.56 with Claude Code,
compared with 69.55 for the corresponding standalone baseline.

\noindent\textbf{Design Harness Track.}
\Cref{tab:controlled-tracks}(a) isolates the design-harness contribution by
holding Claude Code and Claude~4.8 fixed. \texttt{AutoDesign} reaches 74.56, while
Claude Design and OpenDesign score 66.83 and 70.36, respectively, under the
same configuration.

\noindent\textbf{Coding Harness Track.}
\Cref{tab:controlled-tracks}(b) fixes \texttt{AutoDesign} and GLM~5.2, varying only the
coding harness. Kimi Code achieves the highest score at 82.31, followed by
ZCode at 69.53.

\noindent\textbf{Model Track.}
\Cref{tab:controlled-tracks}(c) fixes \texttt{AutoDesign} and Claude Code to separate
model choice from harness variation. Claude~4.8 achieves the highest score at
74.56, followed by Seed~2.1 Pro at 71.83 and Kimi K2.7 at 70.12.

\begin{table}[t]
  \centering
  \caption{Controlled track analysis on \texttt{PosterBench-mini}, the fixed 10-paper subset. (a) varies
  the design harness while holding the coding harness and model fixed. (b)
  varies the coding harness while holding \texttt{AutoDesign} and GLM~5.2 fixed. (c)
  varies the model while holding \texttt{AutoDesign} and Claude Code fixed.}
  \scriptsize
  \setlength{\tabcolsep}{2pt}
  \renewcommand{\arraystretch}{1.12}
  \begin{tabularx}{\textwidth}{@{}
    >{\raggedright\arraybackslash}p{0.19\textwidth}
    *{8}{>{\centering\arraybackslash}X}@{}}
    \toprule
    \textbf{Variant} & \textbf{Overall} $\uparrow$ & \textbf{Faith.} &
    \textbf{Cover.} & \textbf{Density} & \textbf{Vis. Ev.} & \textbf{Layout} &
    \textbf{Read.} & \textbf{Aesth.} \\
    \midrule
    \multicolumn{9}{@{}l}{\textit{\textbf{(a) Design Harness Track}}\quad
      \textit{Coding Harness: Claude Code \;|\; Model: Claude 4.8}} \\
    \addlinespace[2pt]
    \rowcolor[HTML]{E9EEFF}
    \textbf{AutoDesign} & \textbf{74.56} & 9.28 & 8.80 & 8.15 & 5.00 & 9.00 & 7.86 & 6.75 \\
    OpenDesign & 70.36 & 9.10 & 9.00 & 7.39 & 6.15 & 7.13 & 6.38 & 7.31 \\
    Claude Design & 66.83 & 8.95 & 10.00 & 5.24 & 7.90 & 8.54 & 6.09 & 6.77 \\
    \midrule
    \multicolumn{9}{@{}l}{\textit{\textbf{(b) Coding Harness Track}}\quad
      \textit{Design Harness: \texttt{AutoDesign} \;|\; Model: GLM 5.2}} \\
    \addlinespace[2pt]
    \rowcolor[HTML]{E9EEFF}
    \textbf{Kimi Code} & \textbf{82.31} & 9.29 & 9.60 & 7.95 & 6.75 & 8.75 & 8.46 & 7.68 \\
    ZCode & 69.53 & 9.42 & 8.90 & 7.04 & 5.10 & 7.23 & 6.93 & 6.77 \\
    OpenCode & 67.87 & 9.32 & 8.20 & 7.28 & 6.90 & 7.84 & 7.21 & 6.47 \\
    Claude Code & 64.33 & 8.81 & 7.30 & 7.38 & 4.50 & 7.74 & 6.30 & 5.68 \\
    \midrule
    \multicolumn{9}{@{}l}{\textit{\textbf{(c) Model Track}}\quad
      \textit{Design Harness: \texttt{AutoDesign} \;|\; Coding Harness: Claude Code}} \\
    \addlinespace[2pt]
    \rowcolor[HTML]{E9EEFF}
    \textbf{Claude 4.8} & \textbf{74.56} & 9.28 & 8.80 & 8.15 & 5.00 & 9.00 & 7.86 & 6.75 \\
    Seed 2.1 Pro & 71.83 & 9.06 & 9.00 & 6.53 & 6.55 & 9.16 & 7.11 & 6.85 \\
    Kimi K2.7 & 70.12 & 8.54 & 7.00 & 8.30 & 6.27 & 7.81 & 6.82 & 6.00 \\
    GLM 5.2 & 64.33 & 8.81 & 7.30 & 7.38 & 4.50 & 7.74 & 6.30 & 5.68 \\
    LongCat 2.0 & 55.13 & 9.11 & 8.10 & 5.24 & 5.25 & 7.56 & 5.81 & 5.65 \\
    DeepSeek V4 Pro & 54.29 & 9.47 & 9.60 & 4.32 & 5.50 & 8.38 & 5.62 & 6.00 \\
    \bottomrule
  \end{tabularx}
  \label{tab:controlled-tracks}
\end{table}

\subsection{Ablation Studies}

\subsubsection{Effect of the Design Harness}

To isolate the design harness contribution, we hold the model and coding agent
fixed and compare each configuration before and after harness attachment in
\Cref{tab:harness-attachment-gains}. Across the seven completed configurations,
the harness improves the \texttt{PosterBench} Score by 5.01--19.56 points. Under the native
Codex--GPT-5.5 configuration, it raises the score from 75.87 to 81.46 (+5.59);
with Claude Code and Kimi K2.7, it rises from 57.20 to 70.12 (+12.92). The
largest gain is 19.56 points for DeepSeek~V4 Pro with Claude Code. These
improvements span multiple model--code-agent pairs.
MLLMs have an additional repair signal unavailable to text-only LLMs: at each
attempt, the rendered preview from the preceding attempt is supplied as visual
context for the next repair. This lets the model inspect the artifact it is
editing and localize layout, clipping, or visual-evidence failures that are not
fully captured by textual diagnostics alone.

\definecolor{abOpenAI}{HTML}{5579B5}
\definecolor{abClaude}{HTML}{B9785D}
\definecolor{abKimi}{HTML}{7B6AAE}
\definecolor{abGLM}{HTML}{AD6686}
\definecolor{abDeepSeek}{HTML}{5189A2}
\definecolor{abSeed}{HTML}{56877F}
\definecolor{abLongCat}{HTML}{63926B}
\newcommand{\abconfig}[3]{%
  \textcolor{#1}{\rule{1.4mm}{1.4mm}}\hspace{0.3em}
  \makecell[l]{{\footnotesize #2}\\[-2pt]{\scriptsize\color{black!58}#3}}%
}
\newcommand{\abgainbar}[3]{%
  \begin{tikzpicture}[baseline=-0.45ex]
    \fill[black!8] (0,0) rectangle (0.66,0.16);
    \fill[#1] (0,0) rectangle (#2,0.16);
    \node[anchor=west,text=#1,font=\scriptsize\rmfamily\bfseries]
      at (0.74,0.08) {+#3};
  \end{tikzpicture}%
}

\begin{table}[!htbp]
  \centering
  \begin{minipage}[t]{0.76\textwidth}
    \vspace{0pt}
    \centering
    \small
    \setlength{\tabcolsep}{0pt}
    \renewcommand{\arraystretch}{1.13}
    \begin{tabularx}{\linewidth}{@{}
      >{\raggedright\arraybackslash}X
      >{\centering\arraybackslash}p{0.15\linewidth}
      >{\centering\arraybackslash}p{0.21\linewidth}
      >{\centering\arraybackslash}p{0.30\linewidth}@{}}
      \toprule
      \multicolumn{1}{c}{Configuration} & \multicolumn{1}{c}{Original}
        & \multicolumn{1}{c}{\texttt{AutoDesign}} & \multicolumn{1}{c}{Gain} \\
      \midrule
      \abconfig{abOpenAI}{GPT-5.5}{Codex}
        & 75.87 & {\rmfamily\bfseries 81.46} & \abgainbar{abOpenAI}{0.10}{5.59} \\
      \abconfig{abClaude}{Claude 4.8}{Claude Code}
        & 69.55 & {\rmfamily\bfseries 74.56} & \abgainbar{abClaude}{0.09}{5.01} \\
      \abconfig{abSeed}{Seed 2.1 Pro}{Claude Code}
        & 54.01 & {\rmfamily\bfseries 71.83} & \abgainbar{abSeed}{0.33}{17.82} \\
      \abconfig{abKimi}{Kimi K2.7 Code}{Claude Code}
        & 57.20 & {\rmfamily\bfseries 70.12} & \abgainbar{abKimi}{0.24}{12.92} \\
      \abconfig{abGLM}{GLM 5.2}{Claude Code}
        & 50.32 & {\rmfamily\bfseries 64.33} & \abgainbar{abGLM}{0.26}{14.01} \\
      \abconfig{abLongCat}{LongCat 2.0}{Claude Code}
        & 43.26 & {\rmfamily\bfseries 55.13} & \abgainbar{abLongCat}{0.22}{11.87} \\
      \abconfig{abDeepSeek}{DeepSeek V4 Pro}{Claude Code}
        & 34.73 & {\rmfamily\bfseries 54.29} & \abgainbar{abDeepSeek}{0.36}{19.56} \\
      \bottomrule
    \end{tabularx}
  \end{minipage}
  \caption{Effect of attaching \texttt{DesignHarness} while keeping the
  model and coding agent fixed. Exact \texttt{PosterBench} Scores for each completed
  configuration are ordered by final performance.}
  \label{tab:harness-attachment-gains}
\end{table}

\subsubsection{Cost--Performance Trade-off}

\begin{figure}[tbp]
  \centering
  \includegraphics[width=0.72\linewidth]{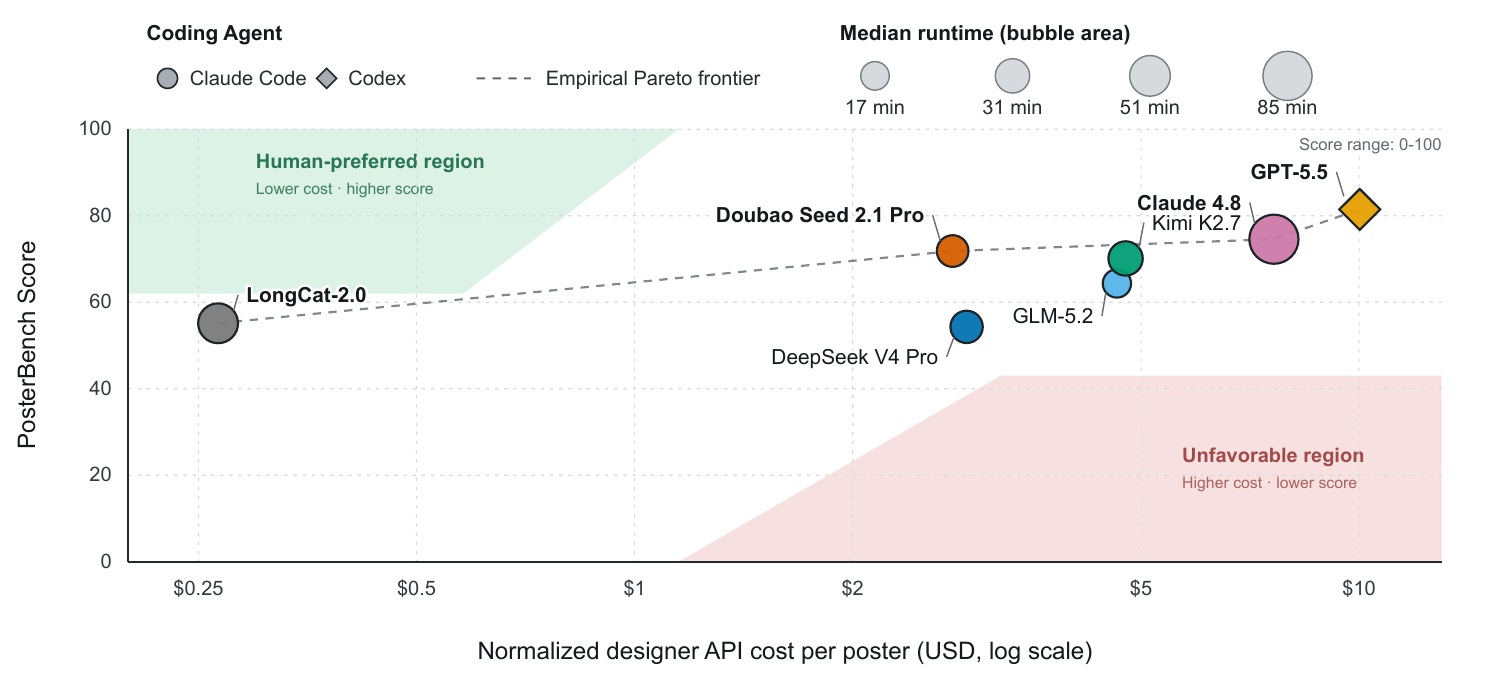}
  \caption{Cost--performance trade-off across seven \texttt{AutoDesign} model
  configurations on \texttt{PosterBench-mini}, the fixed 10-paper subset; marker size denotes median runtime, and the dashed line denotes the empirical Pareto frontier.}
  \label{fig:cost-effectiveness}
\end{figure}

\Cref{fig:cost-effectiveness} compares \texttt{PosterBench} Score with a normalized
designer-only API cost proxy across the seven \texttt{AutoDesign} model configurations
on \texttt{PosterBench-mini}, the fixed 10-paper subset.
The observed Pareto frontier runs from LongCat-2.0\footnote{The reported cost reflects LongCat-2.0's pricing policy at evaluation time: cached context incurs no charge on a cache hit.} (55.13 at \$0.27 per
poster), through Doubao Seed~2.1 Pro (71.83 at \$2.75) and Claude~4.8 (74.56
at \$7.63), to GPT-5.5 (81.46 at \$10.02). Doubao reaches 88\% of the GPT-5.5
score at 27\% of its cost, while GPT-5.5 provides the highest absolute
performance.

\subsection{Human Evaluation}
\label{sec:human-evaluation}

We recruited 11 volunteer reviewers for a fully system-blind pairwise
evaluation of \texttt{AutoDesign}, Claude Code, OpenDesign, and Claude Design on the 100
source papers of the \texttt{PosterBench} Main Track. For each comparison, reviewers saw only two posters generated
from the same paper; no method, system, or model identity was disclosed. They
selected the left poster, the right poster, approximate equality, or skip. We
fit a Bradley--Terry model~\citep{bradley1952rank},
\[
\Pr(i \succ j)
=\frac{\exp(\beta_i)}{\exp(\beta_i)+\exp(\beta_j)},
\]
where $\beta_i$ is the latent preference strength of system $i$. Ties contribute
one half-win to each system and skips are excluded. We report the probability
of beating a uniformly sampled alternative, with 95\% intervals from 2,000
crossed bootstrap resamples of papers and reviewers.
The complete task roster, decision schema, and blind-review interface are
documented in \Cref{app:execution-accounting} and \Cref{fig:human-evaluation-interface}.

\begin{figure}[H]
  \centering
  \includegraphics[width=0.92\textwidth]{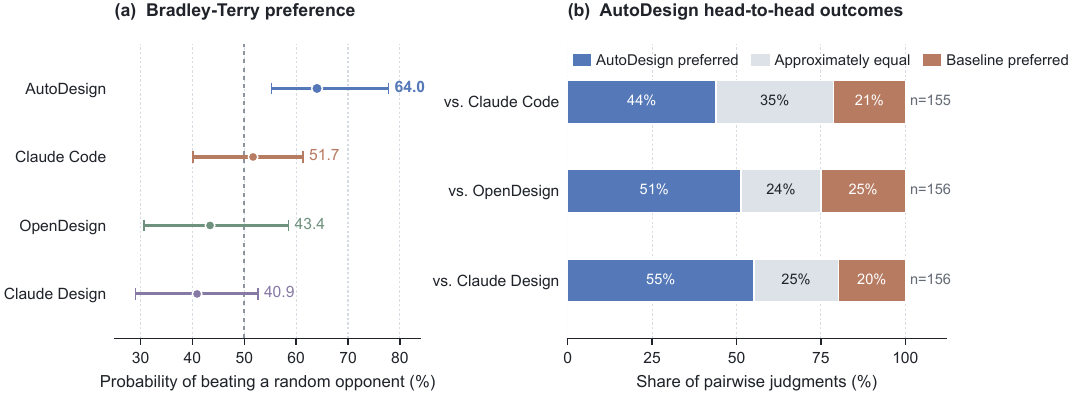}
  \caption{System-blind human evaluation. (a) Bradley--Terry
  probability of beating a uniformly sampled alternative, with 95\% intervals
  crossed by paper--reviewer bootstrap replicates. (b) \texttt{AutoDesign}
  head-to-head outcomes against each baseline. Eleven volunteer reviewers
  submitted 936 responses, including 933 ranking judgments and three
  skips.}
  \label{fig:human-evaluation}
\end{figure}

As shown in \Cref{fig:human-evaluation}, the volunteer reviewers submitted 936
responses: 933 ranking judgments and three skips. \texttt{AutoDesign} has the
highest Bradley--Terry point
estimate at 64.0\% (95\% interval:
55.2--77.8\%). Its tie-adjusted empirical preference scores are 61.3\% against
Claude Code, 63.1\% against OpenDesign, and 67.6\% against Claude Design.

\begin{figure}[H]
  \centering
  \includegraphics[width=0.72\textwidth]{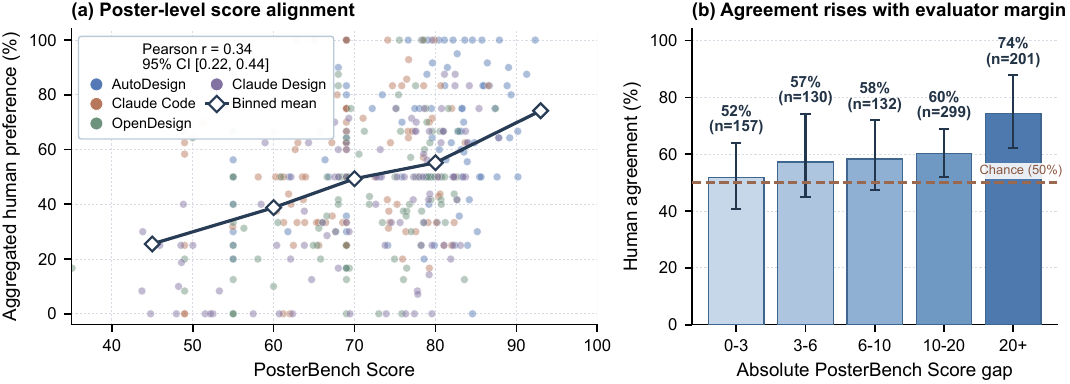}
  \caption{\texttt{PosterBench} Score alignment with system-blind human preference.
  (a) Poster-level score association. (b) Human agreement increases with the
  \texttt{PosterBench} Score margin.}
  \label{fig:benchmark-human-alignment}
\end{figure}

\noindent\textbf{Benchmark--human alignment.}
The blind study also assesses whether \texttt{PosterBench} provides an informative
comparison signal at the poster level. \Cref{fig:benchmark-human-alignment}(a)
compares each poster's \texttt{PosterBench} Score with its tie-adjusted human
preference. Each human score aggregates the blind pairwise judgments involving
that poster. The two measurements have a positive, albeit imperfect,
association ($r=0.34$; a paper-cluster bootstrap gives a 95\% interval of
$[0.22, 0.44]$). This is a useful property of the protocol rather than a
requirement that it duplicate human preference: \texttt{PosterBench} also evaluates
Faithfulness, Coverage, Density, Visual Evidence, Layout, Readability, and
Aesthetics, whereas each blind judgment asks for an immediate pairwise
choice.

More importantly, the \texttt{PosterBench} Score margin calibrates how informative an automatic
comparison is. \Cref{fig:benchmark-human-alignment}(b) uses 919 of the
933 non-skip judgments: the remaining 14 assign equal \texttt{PosterBench} Scores to both
posters and therefore have no benchmark-preferred direction. The probability
that the benchmark-preferred poster agrees with the human decision rises from
51.9\% for 0--3-point gaps to 74.4\% for gaps of at least 20 points. The error
bars are 95\% intervals from 2,000 crossed paper--reviewer bootstrap
replicates, which resample both source papers and reviewers. Thus, the benchmark
offers more than a global system ranking: a large score gap identifies
comparisons in which human preference is substantially more consistent.

\subsection{Qualitative Analysis of Designer's Trajectory}

\begin{figure}[H]
  \centering
  \includegraphics[width=\textwidth]{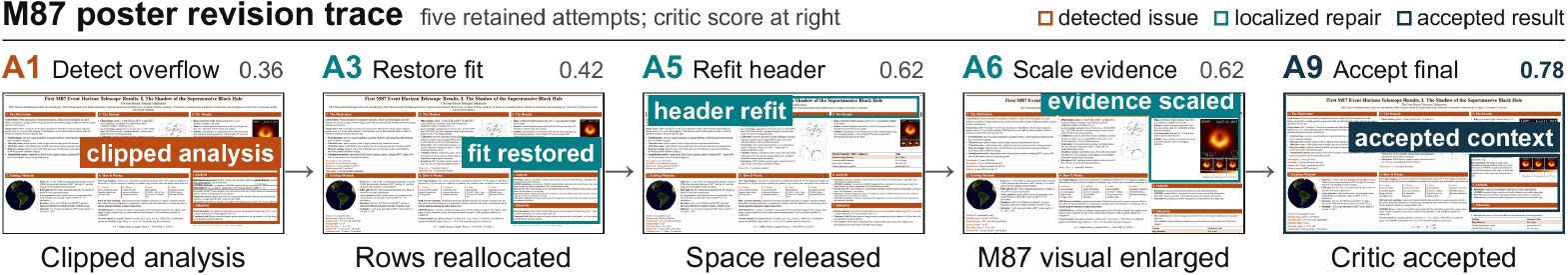}
  \caption{Qualitative trajectory of a poster generation run. Five
  selected attempts show the local change associated with each
  diagnostic or repair event. Colored boxes mark the implicated region: an
  analysis overflow at A1, its fit restoration at A3, header refitting at A5,
  evidence scaling at A6, and the critic-accepted final at A9.}
  \label{fig:qualitative-designer-trajectory}
\end{figure}

\Cref{fig:qualitative-designer-trajectory} traces one 
poster run across five attempts. The critic first identifies
a clipped analysis lane at A1 (0.36); the reallocation of the row removes the constraint at
A3 (0.42), and later the refit of the header and the scaling of evidence produce a more balanced
hierarchy at A5/A6 (0.62). A9 preserves the repaired composition and is accepted
at 0.78. The trace illustrates the intended use of diagnostic feedback: edits
stay localized to the failing region while valid layout and source-derived content are retained across revisions.

\FloatBarrier

\section{Future Directions}
\label{sec:future-directions}

\begin{figure}[H]
  \centering
  \makebox[\textwidth][c]{%
    \includegraphics[width=0.60\textwidth]{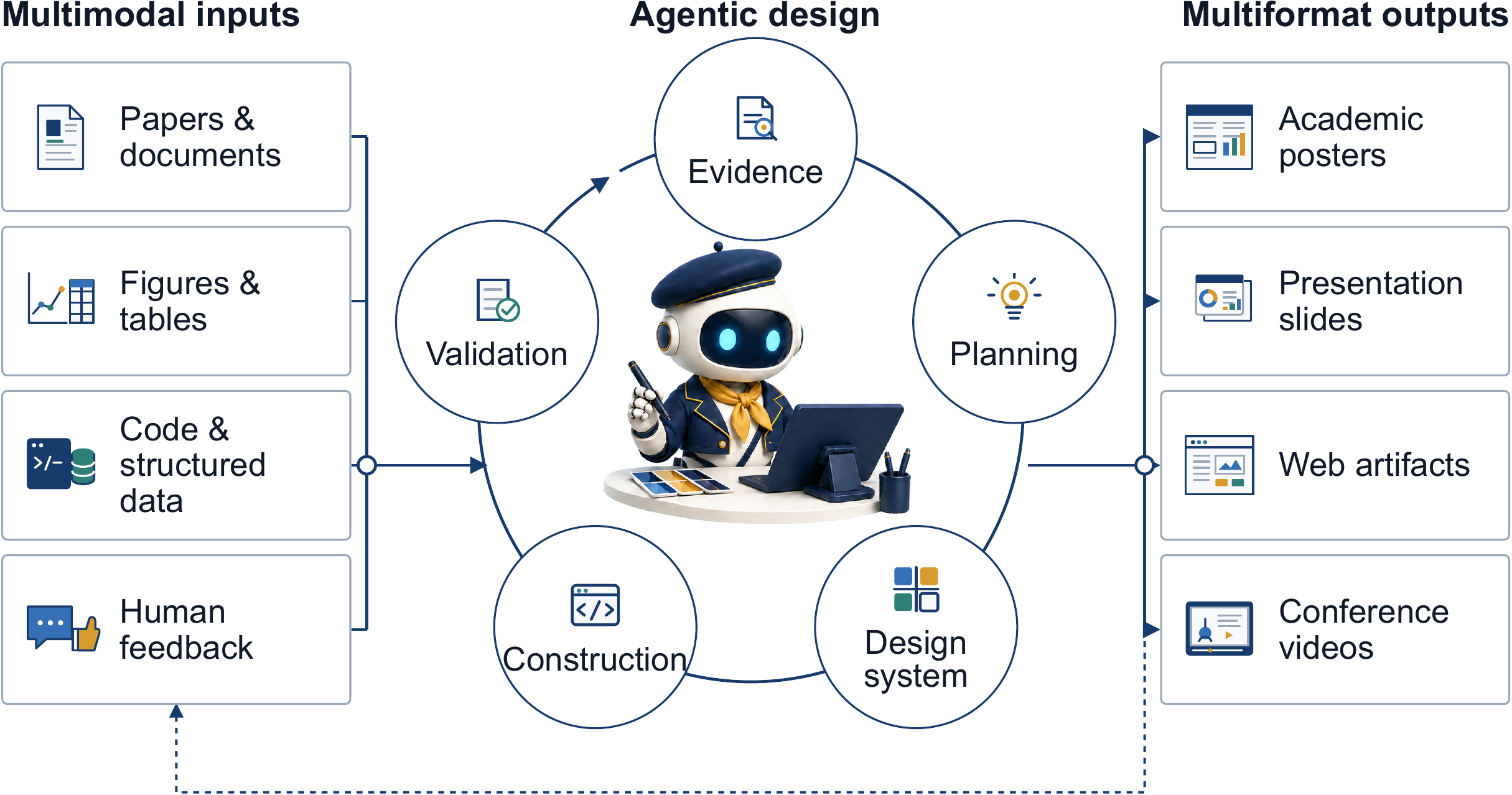}%
  }
  \caption{Future direction: a multimodal-in, multimodal-out agentic design
  system that integrates diverse sources and human guidance to iteratively
  create medium-specific outputs.}
\label{fig:future-mimo}
\end{figure}

\begin{figure}[H]
  \centering
  \makebox[\textwidth][c]{%
    \includegraphics[width=1.0\textwidth]{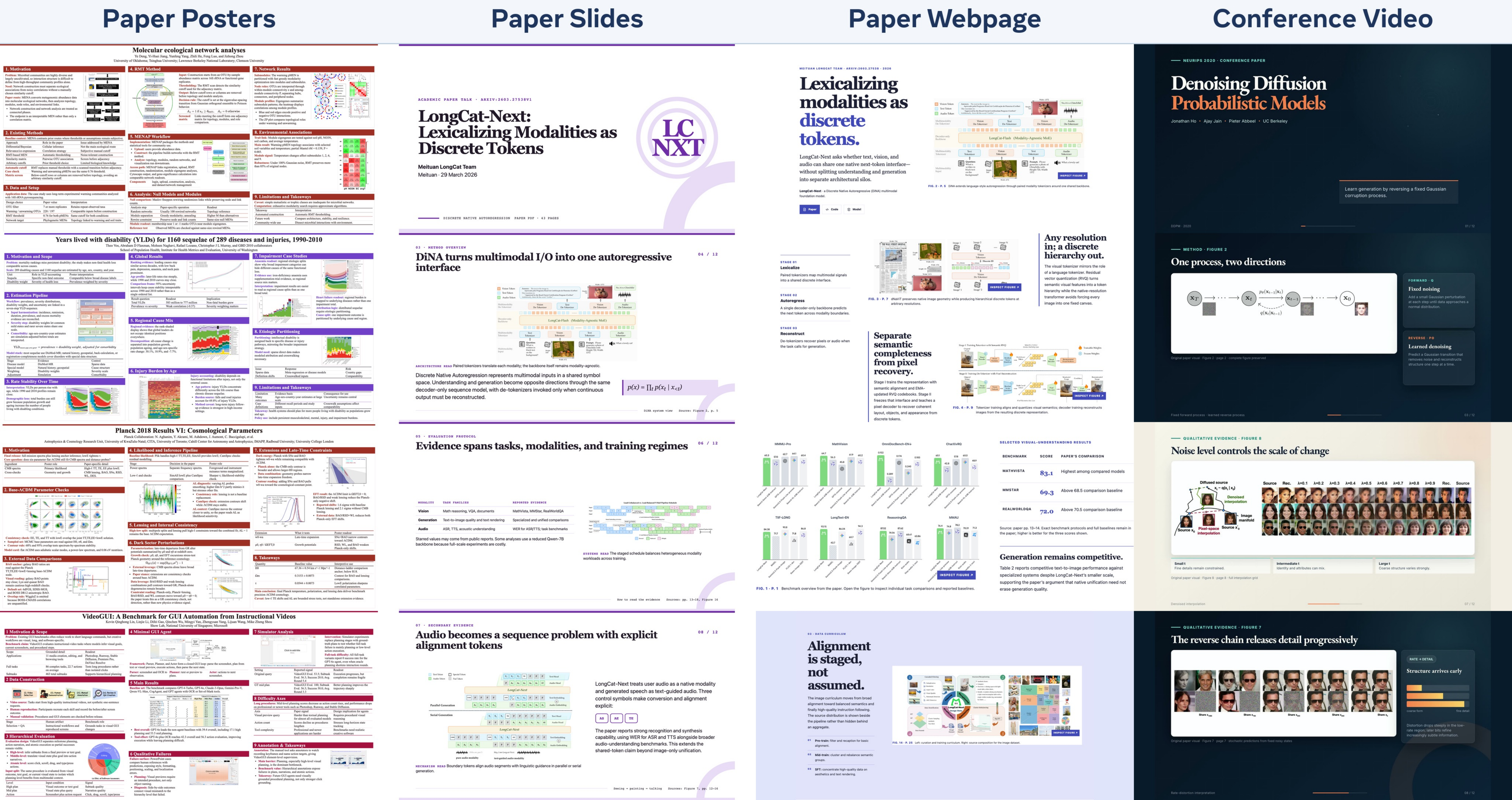}%
  }
  \caption{\textbf{Future directions for AutoDesign.} The initially optimized
    \texttt{DesignHarness} already extends paper-to-poster generation to slide
    decks, webpages, and conference videos. Applying the same meta-harness
    optimization methodology to these and other multimodal outputs offers a path
    toward a general multimodal-in \& out agentic design system.}
  \label{fig:autodesign-multiformat}
\end{figure}

\texttt{AutoDesign} is presently validated for academic paper-to-poster generation, but
the underlying agentic design pattern is not tied to a single input or output
medium. The pilot artifacts in \Cref{fig:autodesign-multiformat} show that the
current \texttt{DesignHarness} can also produce paper-to-slide,
paper-to-webpage, and paper-to-conference-video outputs. \Cref{fig:future-mimo}
illustrates the long-term direction: a multimodal-in, multimodal-out agentic
design system that integrates papers, visual evidence, code, data, and human
guidance, then iteratively creates the medium-appropriate output for a target
communication setting.
Making this expansion reliable requires more than exposing new output formats.
Each medium needs source--output data, an evaluator, a rendering and validation
gate, and an objective tailored to its communication setting. \texttt{PosterBench}
formally evaluates academic posters only; the slide, webpage, and video
artifacts therefore remain pilots. Shared context construction, preference
memories, and repair histories could nevertheless provide a substrate for
reusing experience across media, provided that the transfer is evaluated against
medium-specific objectives.
At  meta-harness level, better component selection and evaluator evolution
remain open problems. A selector should choose  next bounded update from
failure attribution, uncertainty, expected improvement, and component
interactions. Any adaptive evaluator must remain versioned and anchored by
frozen reference tasks, adversarial probes, and periodic human audits so 
optimization doesn't reward-hack a moving target.

Finally, recent work has begun to connect continual harness adaptation,
self-improving harnesses, and model--harness co-evolution
\citep{karten2026continualharness,zhang2026selfharness,lee2026rhi}. In this
direction, harness optimization can complement model post-training: long-horizon
trajectories and repair outcomes provide execution-time supervision, whereas the
model supplies the reasoning and coding capabilities. Joint training should
preserve this division while evaluating both layers against shared held-out
objectives.

\section{Related Work}
\label{sec:related}

Systems for multimodal output generation transform heterogeneous sources into
audience- and medium-specific outputs. For academic posters, SciPostLayout and
deep submodular extraction study layout data, source coverage, and text--image
alignment, while Paper2Poster, P2P, PosterGen, PosterForest, and Any2Poster
combine multimodal inputs with generation, specialized agents, and visual
refinement~\citep{wang2024scipostlayout,jaisankar2025poster,
pang2025paper2poster,sun2025p2p,zhang2025postergen,choi2026posterforest,
vinaykumar2026any2poster}. Closely related systems generate slides, webpages,
and narrated videos from papers or documents~\citep{fu2022doc2ppt,zheng2025pptagent,ge2025autopresent,yang2025autoslides,
chen2025paper2web,zhu2025paper2video}. Structured representations such as
HTML/CSS and explicit layers further support editability, rendering, and visual
inspection~\citep{qu2025igd,liu2026designascode,si2025design2code,wu2024uicoder}.
Together, these works establish the input, output, and representation choices
for multimodal output generation, but their run-time feedback generally remains
within a fixed production procedure.

Critics, render diagnostics, and regeneration policies can improve the current
multimodal output without changing the system that generated it. Self-Refine is
the canonical response-level instance: feedback is used to revise the current
answer~\citep{madaan2023selfrefine}. Other agent methods retain experience more
persistently: Reflexion stores verbal reflections, Voyager accumulates
executable skills, and ExpeL extracts reusable experience from solved tasks~\citep{shinn2023reflexion,wang2023voyager,zhao2024expel}. These mechanisms
preserve useful information beyond a single attempt, but they typically do not
update the harness that repeatedly produces outputs.
This system-level perspective has early roots in classical accounts of
autonomous agents, self-referential learning, and retaining policy changes
according to long-term reward effects~\citep{wooldridge1995intelligentagents,
schmidhuber1987selfreferential,schmidhuber1997successstory}. These works are
conceptual precedents rather than direct algorithms for \texttt{AutoDesign}.
Recent work separates improvement of model parameters from persistent
improvement of the operational scaffold around a fixed
model~\citep{ren2026selfimprovement}. TextGrad, DSPy, and GEPA optimize
components or declarative pipelines, whereas STOP, GPTSwarm, ADAS, and AFlow
search code- or graph-represented workflows~\citep{yuksekgonul2024textgrad,khattab2024dspy,agrawal2025gepa,
zelikman2024stop,zhuge2024gptswarm,hu2025adas,zhang2024aflow}. At the
full-harness level, A Self-Improving Coding Agent and MOSS update agent source
from execution evidence; Meta-Harness, HarnessX, Self-Harness, and Agentic
Harness Engineering study searchable harness programs, composable primitives,
bounded updates, and outcome attribution~\citep{robeyns2025selfimproving,cai2026moss,
lee2026metaharness,chen2026harnessx,zhang2026selfharness,lin2026ahe}. HarnessX
also uses execution traces as signals for both harness evolution and future
model training~\citep{chen2026harnessx}. Recursive Harness
Self-Improvement specializes a user-constructed, prompt-level multi-agent
harness from pairwise revision feedback~\citep{lee2026rhi}. G\"odel Machines
provide a proof-based ideal for self-rewriting, while Darwin G\"odel Machine
and Huxley-G\"odel Machine study empirical harness evolution and distinguish
immediate performance from future self-improvement
potential~\citep{schmidhuber2006godelmachine,zhang2026dgm,wang2025huxley}.
\texttt{AutoDesign} instantiates this direction for academic design by evolving
a \texttt{DesignHarness} for source-grounded, editable artifacts.

Persistent system updates also differ in how they separate learning signals from
final evaluation. RHI~\citep{lee2026rhi} keeps its evaluation prompt on the evaluator
side of the update loop, yet the resulting pairwise history remains a task-local
learning signal. Recursive Self-Evolving Agents use an
independent development split to gate persistent updates, while Continual
Harness, Adaptive Auto-Harness, and Live-SWE-agent study related adaptation
settings~\citep{nguyen2026rsea,karten2026continualharness,
liu2026adaptiveautoharness,xia2025livesweagent}. Agent-as-a-Judge further
highlights the value of process-level evidence alongside final-outcome
assessment~\citep{zhuge2025agentasjudge}. \texttt{AutoDesign} uses an independent
development acceptance gate for harness updates and evaluates the final
\texttt{DesignHarness} with \texttt{PosterBench} and a system-blind human study.

\section{Conclusion}
\label{sec:conclude}

\texttt{AutoDesign} turns recurring design failures into improvements to the system that generates future multimodal outputs. Its \texttt{MetaHarnessOptimizer} aggregates trajectories, source and rendering diagnostics, evaluator feedback, and reference posters to update one \texttt{DesignHarness} component at a time while keeping model weights fixed. This makes paper-to-poster generation a persistent learning process that accumulates design priors and produces editable outputs for direct use or local revision on the \href{https://designanything.ai/}{Demo Page}.
We also introduce \texttt{PosterBench}, a unified evaluation protocol for academic posters. \texttt{DesignHarness} achieves the top score (78.32) and, with Claude Code and Claude~4.8 fixed, surpasses Claude Design by 7.45 points. It also receives the highest Bradley--Terry estimate in a system-blind human study.

\bibliographystyle{assets/plainnat}
\bibliography{resources/main}

\clearpage
\appendix
\section{Supplementary Experimental Materials}
\label{sec:appendix}

This appendix makes the experimental interface inspectable: it documents the
controlled comparisons, shared generation instruction, frozen scoring protocol,
released records, and matched visual evidence that complement the main paper.

\begin{appendixrecordbox}{Appendix guide}
\small
\setlength{\tabcolsep}{2.5pt}
\renewcommand{\arraystretch}{1.18}
\begin{tabularx}{\columnwidth}{@{}>{\raggedright\arraybackslash}p{0.37\columnwidth}X@{}}
\toprule
\textbf{Jump to} & \textbf{Contents} \\
\midrule
\hyperref[app:benchmark-inputs]{\textbf{A.1 Benchmark Inputs and Comparison Matrix}} &
Source packages, the factors held fixed in each comparison, and the factor varied by every reported track. \\
\hyperref[app:shared-generation-prompt]{\textbf{A.2 AutoDesign Generation Interface}} &
The system-instruction excerpt and the shared task prompt used across compared systems. \\
\hyperref[app:design-harness-evolution]{\textbf{A.3 DesignHarness Evolution}} &
The staged capabilities accumulated in the optimized design harness and their mapping to the five-component abstraction. \\
\hyperref[app:poster-evaluation-criteria]{\textbf{A.4 PosterBench Evaluation Interface}} &
The seven-dimension rubric, score aggregation, protected gates, and per-case evaluation record. \\
\hyperref[app:execution-accounting]{\textbf{A.5 Released Records and Human Evaluation}} &
The benchmark archive schema, blind-review interface, and system-blind human-evaluation protocol. \\
\hyperref[fig:appendix-qualitative-matrix]{\textbf{Matched Qualitative Comparisons}} &
Curated three-system poster comparisons on identical source papers. \\
\hyperref[fig:appendix-poster-demonstrations]{\textbf{Additional Poster Demonstrations}} &
Four source-grounded, editable paper posters generated by \texttt{AutoDesign}. \\
\bottomrule
\end{tabularx}
\end{appendixrecordbox}

\subsection{Benchmark Inputs and Comparison Matrix}
\label{app:benchmark-inputs}

Each case provides the source PDF and available paper assets.  All systems in a
matched row receive the same package and produce one editable poster artifact;
the source paper remains the authority for claims, numbers, and visual evidence.
Table~\ref{tab:appendix-tracks} identifies the controlled factor for every
reported track.

\begin{center}
  \small
  \setlength{\tabcolsep}{4pt}
  \renewcommand{\arraystretch}{1.18}
  \begin{tabularx}{\textwidth}{@{}
    >{\raggedright\arraybackslash}p{0.24\textwidth}
    >{\centering\arraybackslash}p{0.06\textwidth}
    >{\raggedright\arraybackslash}p{0.42\textwidth}
    >{\raggedright\arraybackslash}X@{}}
    \toprule
    \textbf{Track} & \textbf{Papers} & \textbf{Fixed across a row comparison} & \textbf{Varied factor} \\
    \midrule
    \texttt{PosterBench} Main Track & 100 & Source paper, source assets, output contract, and frozen \texttt{PosterBench} protocol & System configuration \\
    \texttt{PosterBench-mini} Main Track & 10 & Shared \texttt{PosterBench-mini} papers, output contract, and frozen \texttt{PosterBench} protocol & System configuration \\
    Design Harness Track & 10 & Claude Code and Claude~4.8; shared \texttt{PosterBench-mini} papers and frozen \texttt{PosterBench} protocol & Design harness \\
    Coding Harness Track & 10 & \texttt{AutoDesign}, GLM~5.2, shared \texttt{PosterBench-mini} papers, and frozen \texttt{PosterBench} protocol & Coding harness \\
    Model Track & 10 & \texttt{AutoDesign}, Claude Code, shared \texttt{PosterBench-mini} papers, and frozen \texttt{PosterBench} protocol & Model \\
    Harness-Attachment Ablation & 10 & Model, code agent, shared \texttt{PosterBench-mini} papers, and frozen \texttt{PosterBench} protocol & Presence of the \texttt{AutoDesign} design harness \\
    \bottomrule
  \end{tabularx}
  \captionof{table}{Controlled comparison matrix.  Main tracks compare complete systems;
  controlled tracks vary only the factor named in the final column.}
  \label{tab:appendix-tracks}
\end{center}

\paragraph{Runtime Versions and Reasoning Configuration.}
Under our controlled configurations, we used \texttt{codex-cli}~v0.142.3 for
Codex and Claude Code~v2.1.119 for Claude Code. Each model used its highest
available thinking-effort setting.

\FloatBarrier

\subsection{AutoDesign Generation Interface}
\label{app:shared-generation-prompt}

The \texttt{AutoDesign} system excerpt is shown first. The shared user prompt below fixes
the target artifact, source-grounding requirements, and visible quality
constraints for every compared system. System-specific orchestration, tools, and
repair policies remain part of each system configuration.

\begin{appendixpromptbox}{AutoDesign System Prompt (excerpt)}
You are the \texttt{AutoDesign} designer. Turn the supplied brief into editable visual
artifacts. Keep final text native and editable. Ground paper-derived claims in
the supplied source; do not invent numbers, authors, venues, URLs, benchmark
deltas, compute, or citations. For an academic paper poster, use the source
figures and tables as primary visual evidence, author a complete editable
artifact, render it, and repair objective layout or grounding defects before
delivery.
\end{appendixpromptbox}

\tcbinputlisting{
  enhanced,
  breakable,
  colback=metabg!65!white,
  colframe=metablue!72!black,
  colbacktitle=metablue!14!white,
  coltitle=metafg,
  fonttitle=\small\sffamily\bfseries,
  title={Shared User Prompt (verbatim)},
  title after break={Shared User Prompt (continued)},
  boxrule=0.5pt,
  arc=3pt,
  left=5pt,
  right=5pt,
  top=4pt,
  bottom=4pt,
  listing only,
  listing engine=listings,
  listing file={resources/benchmark_generation_user_prompt.txt},
  listing options={basicstyle=\footnotesize\ttfamily,breaklines=true,columns=fullflexible}
}

\subsection{DesignHarness Evolution}
\label{app:design-harness-evolution}

\begin{figure}[tbp]
  \centering
  \includegraphics[width=\textwidth]{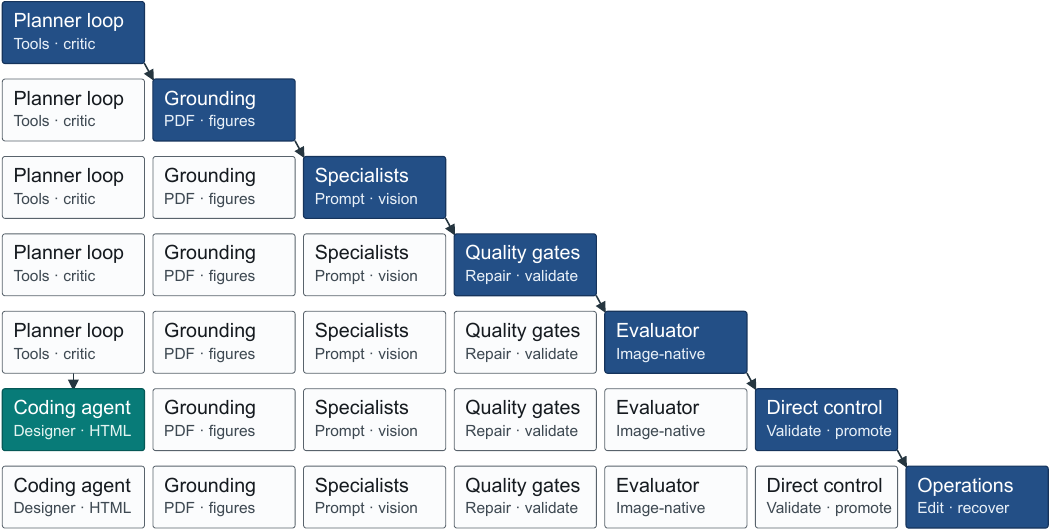}
  \caption{\textbf{DesignHarness architecture evolution for paper-to-poster.}
  An implementation-level summary of accumulated capabilities, not a second
  harness taxonomy or a chronological outer-loop trace. The diagonal path
  highlights additions in this architectural view; it does not denote individual
  meta-harness iterations. The final harness combines source grounding,
  specialist support, quality gates, image-native evaluation, controlled
  promotion, and operations for editable HTML artifacts.}
  \label{fig:appendix-design-harness-evolution}
\end{figure}

\Cref{fig:appendix-design-harness-evolution} provides an implementation-level
architecture summary for the paper-to-poster instantiation of
\texttt{DesignHarness}. It is neither a second taxonomy nor a record of
individual outer-loop iterations: its stages collectively instantiate the five
functional components of $H$ in \Cref{sec:design-harness-definition}. The
illustration moves from a planning-and-critique loop to a coding-agent designer
that authors editable HTML, while source grounding, specialist prompt and vision
support, repair and validation gates, image-native evaluation, candidate
promotion, and recovery operations accumulate around that core.

The figure shows why the optimized result is more than a static prompt or a
single repair loop: it accumulates a coherent system around the fixed model. In
the five-component abstraction, grounding supplies context and memory; specialist
support and editable HTML define tools and specifications; the workspace,
browser, renderer, and export environment support the coding-agent authoring
path as the execution runtime; direct control and operations implement
orchestration; and quality gates together with image-native evaluation provide
feedback for revision. The image-native evaluator shown here belongs to
evaluation and feedback inside the design harness. It is distinct from the
outer-loop evaluator $R_{\mathrm{meta}}$ and from the frozen \texttt{PosterBench} protocol
used for final comparison (\Cref{sec:outer-loop,app:poster-evaluation-criteria}).

\subsection{PosterBench Evaluation Interface}
\label{app:poster-evaluation-criteria}

\texttt{PosterBench} evaluates a rendered artifact against its source paper with seven
scores $q_j \in [0,10]$.  Its rubric, weights, and protected gates are manually
specified and frozen before comparative evaluation.  The seven dimensions use
the same quality vocabulary as the outer-loop evaluator $R_{\mathrm{meta}}$
in Section~\ref{sec:outer-loop}, but serve a different role: \texttt{PosterBench} reports
completed-system quality, whereas an evaluator coding agent constructs
$R_{\mathrm{meta}}$ from annotated reference artifacts to provide update
feedback during harness optimization.  $R_{\mathrm{meta}}$ is then fixed within
each optimization run.

\begin{center}
  \footnotesize
  \setlength{\tabcolsep}{2.5pt}
  \renewcommand{\arraystretch}{1.16}
  \begin{tabularx}{\columnwidth}{@{}
    >{\raggedright\arraybackslash}p{0.17\columnwidth}
    >{\centering\arraybackslash}p{0.07\columnwidth}
    >{\raggedright\arraybackslash}p{0.15\columnwidth}
    >{\raggedright\arraybackslash}X@{}}
    \toprule
    \textbf{Dimension} & \textbf{Wt.} & \textbf{Score mode} & \textbf{Operational definition} \\
    \midrule
    Faithfulness & 10 & Programmatic + VLM & Checks numeric and source grounding, then judges whether claims, entities, and visual evidence remain consistent with the paper. \\
    Coverage & 10 & VLM & Assesses whether the poster preserves the paper's problem, method, evidence, and takeaway against a compact source brief. \\
    Density & 15 & Programmatic & Measures information occupancy, OCR text coverage, blank interiors, and pasted paper-body screenshots. \\
    Visual Evidence & 10 & Programmatic + VLM & Judges whether figures and tables are relevant, readable, and explained locally; guards reject raw paper-body crops. \\
    Layout & 20 & Programmatic & Audits render size and aspect, OCR fallback, clipping, overlap, export-edge damage, and visible placeholders. \\
    Readability & 25 & Programmatic + VLM & Combines poster-scale text and spatial checks with hierarchy, scan-path, balance, and crowding judgments. \\
    Aesthetics & 10 & VLM & Rates academic visual craft, including typography, palette discipline, and compositional coherence. \\
    \bottomrule
  \end{tabularx}
  \captionof{table}{\texttt{PosterBench} scoring protocol. ``Programmatic'' denotes
  image-native and source-grounded checks; ``VLM'' denotes a
  dimension-specific judgment conditioned on the rendered poster and compact
  source context. The weights sum to 100.}
  \label{tab:appendix-rubric}
\end{center}

For record $i$, \texttt{PosterBench} first forms the fixed weighted score from the seven
dimension scores, applies the record-level ceiling $C_i$, and finally averages
capped poster scores across the benchmark. Its ceiling
families address severe layout damage, insufficient presentation viability,
confirmed visible failures, and protected render-integrity gates; inactive
ceiling families take value 100. A standard P0 gate has a ceiling of 40, while
more severe gate types may impose a lower ceiling. Thus, if
$\overline{\mathbf q}=N^{-1}\sum_i\mathbf q_i$ is the vector displayed by a table row,
$\frac{1}{10}\boldsymbol{\alpha}^{\top}\overline{\mathbf q}$ does not
generally equal \textbf{Overall}: the ceiling is applied before the benchmark
average. For batches of at least 20 readable posters, a blinded
style-homogeneity check may only reduce the professional-aesthetics score; it
is not applied to \texttt{PosterBench-mini}, whose 10-poster scale is below that threshold.

The released benchmark rows score rendered poster images. Programmatic signals
cover render integrity, occupancy, OCR readability, and source grounding. Each
VLM judgment receives the rendered image, a compact paper brief, and selected
grounding signals, but no system identity or generation prompt.

\begin{appendixrecordbox}{Per-case evaluation record (schema excerpt)}
\ttfamily\footnotesize
\{\par
\quad "case\_id": "2017-attention-is-all-you-need",\par
\quad "system": "\texttt{AutoDesign} + Claude Code + Claude 4.8",\par
\quad "overall\_score": "<aggregate-score>",\par
\quad "dimension\_scores": \{ "faithfulness": "<0--10>", \ldots \},\par
\quad "evaluation\_status": "scored"\par
\}
\end{appendixrecordbox}

Released records identify the system configuration, source case, evaluation
status, aggregate score, and dimension scores, enabling each reported row to be
audited and reaggregated.

\subsection{Released Records and Human Evaluation}
\label{app:execution-accounting}

The released score archive records system identity, source case and discipline,
evaluation status, aggregate score, and per-dimension scores. It distinguishes
fresh evaluations from reaggregated records and supports independent auditing of
the reported tables.

\begin{appendixrecordbox}{Released benchmark record fields}
\begin{tabularx}{\columnwidth}{@{}>{\bfseries\sffamily}p{0.31\columnwidth}X@{}}
Configuration & System, design harness, coding harness, and model. \\
Source case & Paper identifier and discipline. \\
Evaluation & Status, aggregate score, and seven dimension scores. \\
\end{tabularx}
\end{appendixrecordbox}

\begin{appendixrecordbox}{Human-evaluation record}
\begin{tabularx}{\columnwidth}{@{}>{\bfseries\sffamily}p{0.40\columnwidth}X@{}}
Included reviewer accounts & 11 \\
Submitted responses & 936 (933 rankings; 3 skips) \\
Judgment design & System-blind pairwise comparison on a shared paper \\
Complete task roster & 600 tasks per reviewer, covering all paper--system pairs \\
Ranking analysis & Bradley--Terry; ties contribute one half-win \\
Uncertainty & 2,000 crossed paper--reviewer bootstrap resamples \\
Agreement diagnostic & Nominal Krippendorff coefficient: 0.101 \\
\end{tabularx}
\end{appendixrecordbox}

\begin{figure}[tbp]
  \centering
  \includegraphics[width=1.0\textwidth]{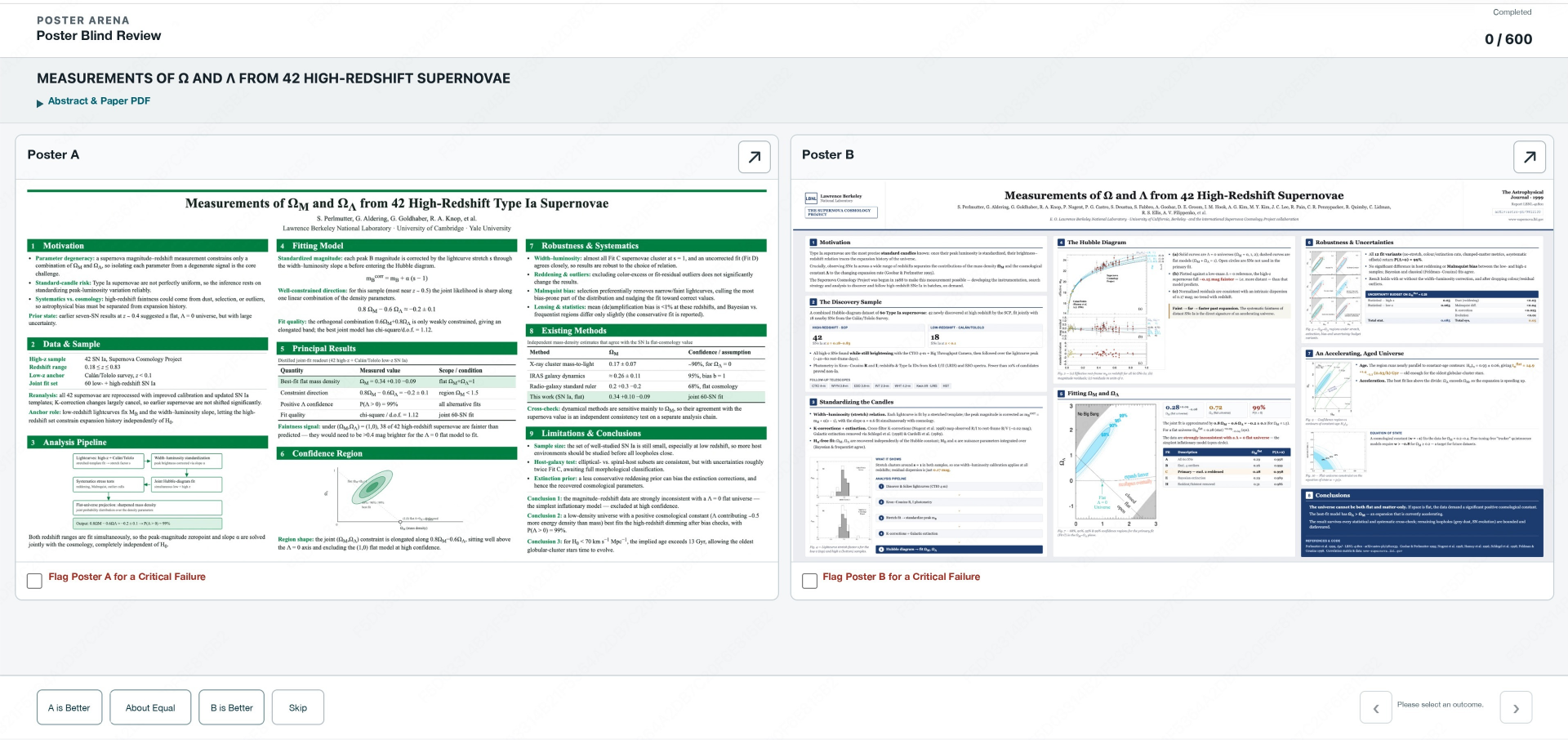}
  \caption{System-blind Poster Arena interface. Each task presents two
  anonymous posters for the same paper, with the abstract and PDF as common
  context. Reviewers select a preference, approximate equality, or skip, and
  can flag a critical failure. The ``0/600'' counter denotes the complete
  roster, rather than a requirement to submit 600 judgments.}
  \label{fig:human-evaluation-interface}
\end{figure}

\paragraph{Blind-review interface and complete task roster.}
\Cref{fig:human-evaluation-interface} shows the website used for the
system-blind study. It presents two anonymous posters for the same source paper
with a shared title, abstract, and PDF; system, model, and harness identities
are withheld. The complete roster balances left--right presentation and records
Poster~A, about equal, Poster~B, skip, and an optional critical-failure flag.

The roster visible to each reviewer contains every paper--system-pair task:
\[
  |\mathcal{T}| = N_{\mathrm{papers}}\binom{N_{\mathrm{systems}}}{2}
  = 100\binom{4}{2} = 600.
\]
It evaluates each of the six unordered system pairs on every paper, providing
equal paper coverage and a connected graph for Bradley--Terry estimation. The
``0/600'' counter denotes this complete roster; submitted non-skip decisions
are retained and uncompleted assignments are not imputed.

\begin{appendixrecordbox}{Human review decision record (schema excerpt)}
\ttfamily\footnotesize
\{\par
\quad "paper\_id": "2017-attention-is-all-you-need",\par
\quad "poster\_a\_id": "anonymous-artifact-a",\par
\quad "poster\_b\_id": "anonymous-artifact-b",\par
\quad "choice": "A | B | approximately-equal | skip",\par
\quad "severe\_a": false,\quad "severe\_b": false,\quad "skip\_reason": null\par
\}
\end{appendixrecordbox}

The released decision record preserves each anonymous comparison and
critical-failure flag. The coefficient is a nominal agreement diagnostic over
paper--system-pair items, not the ranking estimator;
Figure~\ref{fig:benchmark-human-alignment} reports a separate paper-cluster
bootstrap for benchmark--human alignment.

\begin{figure*}[p]
  \centering
  \includegraphics[width=\textwidth,height=0.87\textheight,keepaspectratio]{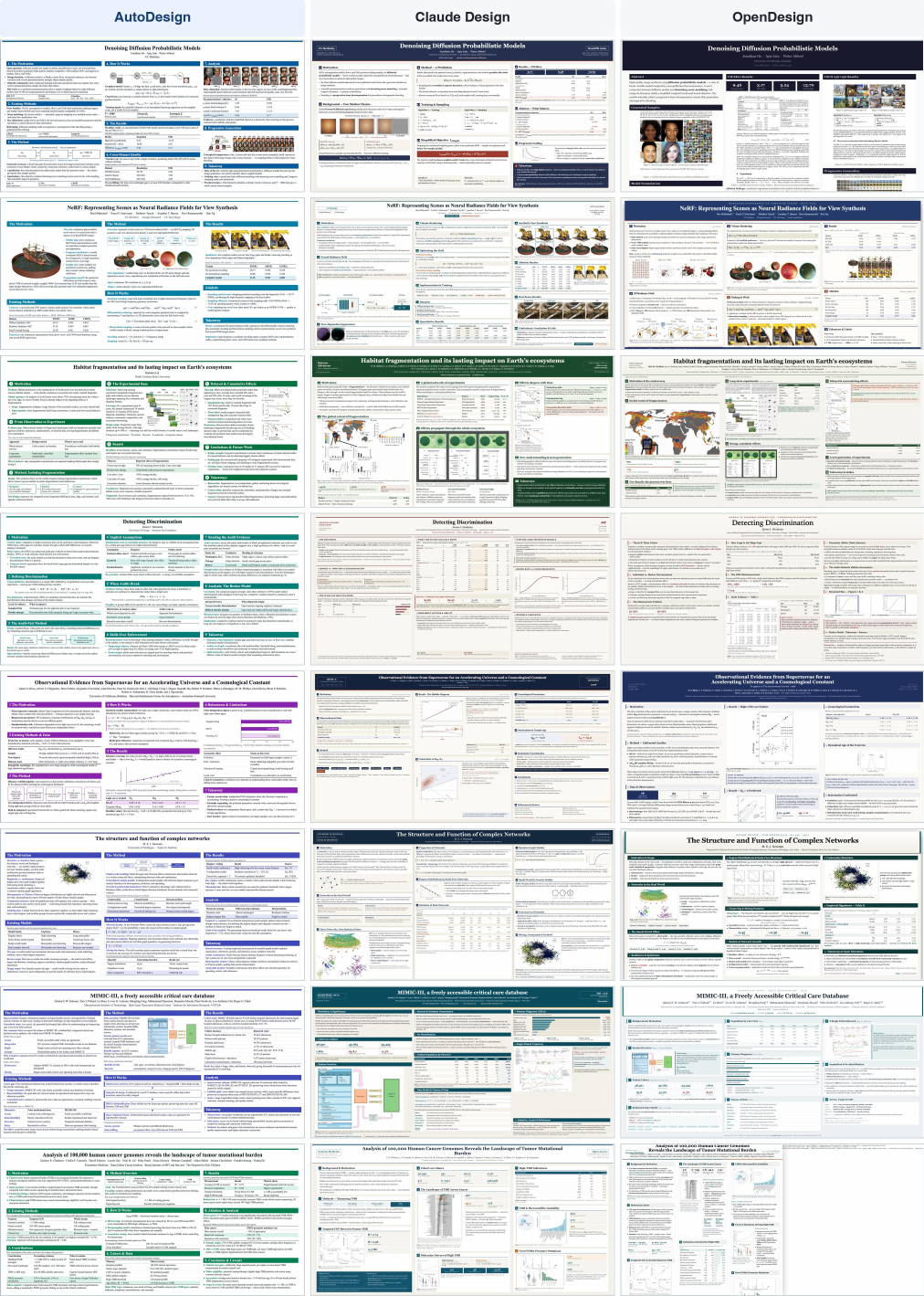}
  \caption{Additional matched qualitative comparisons across Design Agents with the same generation prompt; each row shows three
  independently generated posters for one source paper, at a common scale.}
  \label{fig:appendix-qualitative-matrix}
\end{figure*}

\clearpage
\subsection{Additional AutoDesign Poster Demonstrations}
\label{app:additional-poster-demonstrations}

\begin{figure}[H]
  \centering
  \includegraphics[page=1,width=1.0\linewidth]{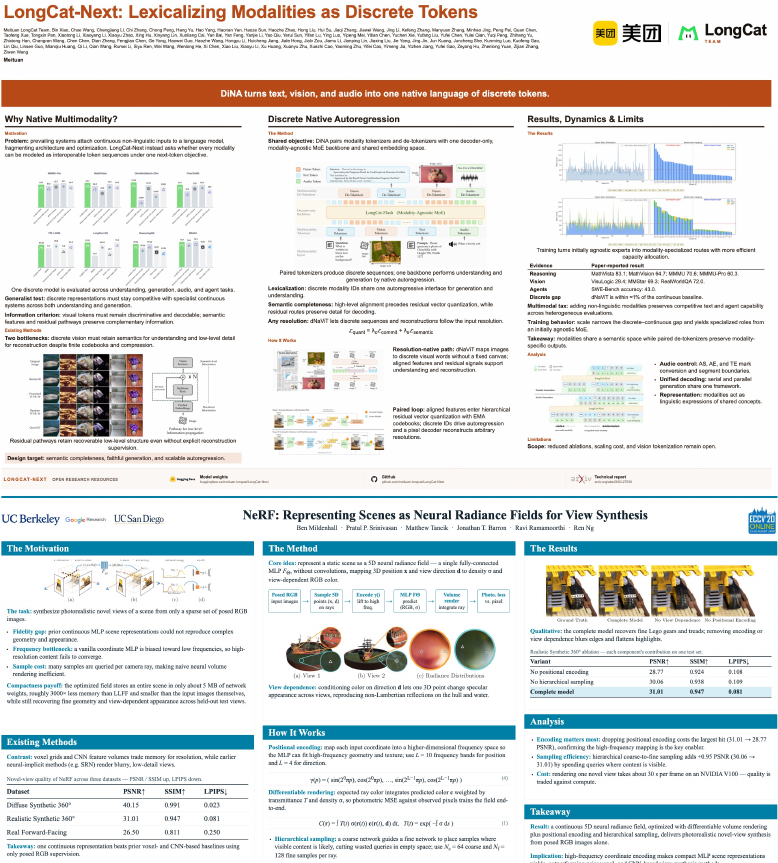}
  \caption{\textbf{Additional AutoDesign poster demonstrations.} LongCat-Next and NeRF
  are rendered from their respective source papers.}
  \label{fig:appendix-poster-demonstrations}
\end{figure}
\clearpage

\begin{figure}[H]
  \centering
  \includegraphics[page=2,width=1.0\linewidth]{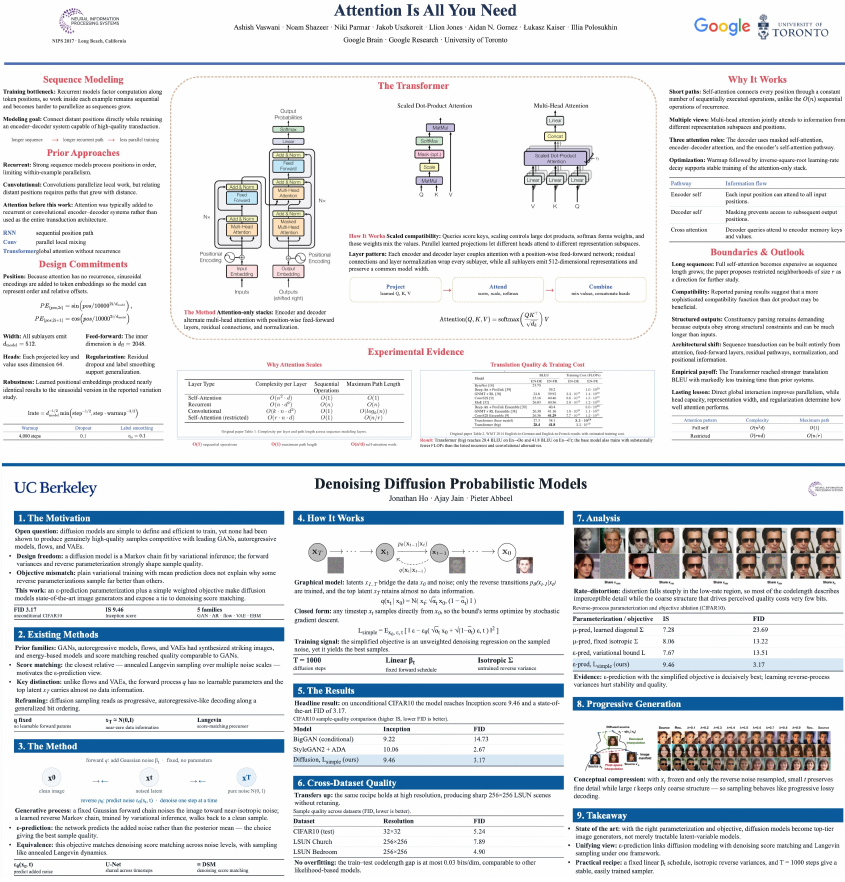}
  \caption{\textbf{Additional AutoDesign poster demonstrations (continued).} Attention Is
  All You Need and DDPM are rendered from their respective source papers.}
\end{figure}
\clearpage

\FloatBarrier

\end{document}